\documentclass[sageh,Afour,times,doublespace]{sagej}

\usepackage{moreverb,url}

\usepackage{subcaption}

\usepackage[colorlinks,bookmarksopen,bookmarksnumbered,citecolor=red,urlcolor=red]{hyperref}

\newcommand\BibTeX{{\rmfamily B\kern-.05em \textsc{i\kern-.025em b}\kern-.08em
T\kern-.1667em\lower.7ex\hbox{E}\kern-.125emX}}

\def\volumeyear{2019}

\begin{document}

\runninghead{Wigness et al.}

\title{Robot navigation from human demonstration: learning control behaviors with environment feature maps}
\author{Maggie Wigness\affilnum{1}, John G. Rogers III\affilnum{1}, Luis E. Navarro-Serment\affilnum{2}}

\affiliation{\affilnum{1}US Army Research Laboratory, USA\\
\affilnum{2}The Robotics Institute, Carnegie Mellon University, USA}

\corrauth{Maggie Wigness,
US Army Research Laboratory,
Adelphi, MD, USA.}

\email{maggie.b.wigness.civ@mail.mil}

\begin{abstract}
When working alongside human collaborators in dynamic and unstructured environments, such as disaster recovery or military operation, fast field adaptation is necessary for an unmanned ground vehicle (UGV) to perform its duties or learn novel tasks. In these scenarios, personnel and equipment are constrained, making training with minimal human supervision a desirable learning attribute. We address the problem of making UGVs more reliable and adaptable teammates with a novel framework that uses visual perception and inverse optimal control to learn traversal costs for environment features. Through extensive evaluation in a real-world environment, we show that our framework requires few human demonstrated trajectory exemplars to learn feature costs that reliably encode several different traversal behaviors. Additionally, we present an on-line version of the framework that allows a human teammate to intervene during live operation to correct deteriorated behavior or to adapt behavior to dynamic changes in complex and unstructured environments.
\end{abstract}


\maketitle

\section{Introduction}

Response teams sent to areas stricken by disaster work in highly dynamic and unstructured environments; every second counts when lives are in danger. Robots have been deployed in these humanitarian assistance and disaster relief (HA/DR) efforts~\citep{murphy2014disaster}, but a number of gaps still need to be addressed to further improve the reliability of these fielded autonomous assets. We focus specifically on the importance of quickly learning and adapting navigational behavior for unmanned ground vehicles (UGVs).

Many high level tasks performed by UGVs under dynamic and unstructured scenarios require autonomous navigation. Advances in visual perception~\citep{simonyan2015very,szegedy2015going,krizhevsky2012imagenet} make it possible to provide fine-grained semantic segmentations~\citep{badrinarayanan2015segnet,long2015fully} of an environment that an UGV can use to make navigation decisions. This includes determining the traversability of terrain~\citep{shneier2008learning,papadakis2013terrain,roncancio2014traversability}. However, in the wake of a disaster, many expected contextual cues for navigation will be unavailable, unreliable, or misleading. For example, visual appearance and traversability of roads may have changed due to being covered in mud or sand after a hurricane. Thus, previously learned behavior may not be relevant and could be dangerous for the UGV to execute. To operate successfully in these scenarios, an UGV will need to be adapted quickly and with minimal human supervision to minimize impact on personnel and equipment.  

\begin{figure*}
\centering
\includegraphics[width=0.85\linewidth]{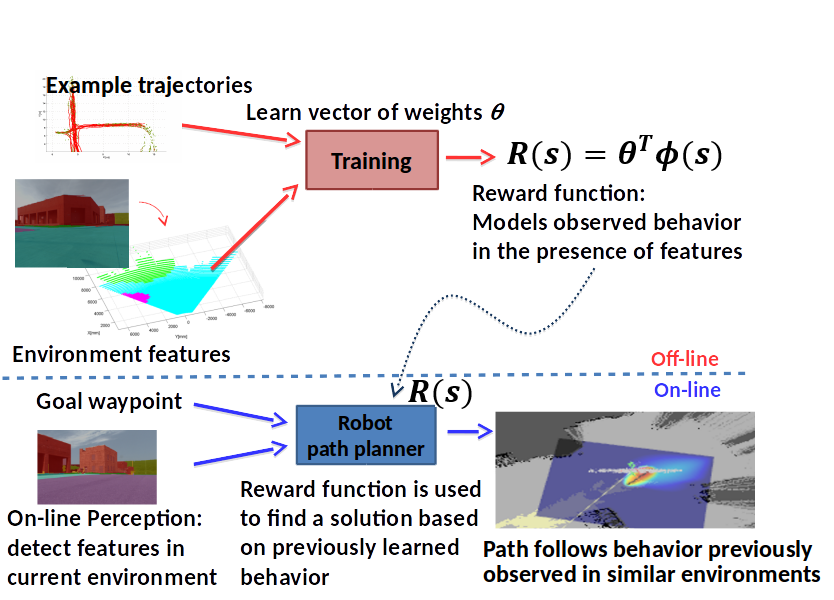}
\caption{Overview of the off-line training process and on-line path planning of our system that learns to assign cost values to regions in an urban environment.}
\label{fig:overview}
\end{figure*}

In addition to the perceptual and control challenges of adapting to support operation in an environment recovering from a disaster, an UGV might be called upon to perform unforeseen and unexpected roles in the field. The dynamic and unpredictable nature of a disaster scenario precludes preparing for every possible role. Therefore, the ability to rapidly train UGVs to perform novel tasks in the field increases their utility to first responders.

Early robot navigation efforts focused on finding minimum distance collision-free paths to a goal. Further developments began considering safety because navigating optimally with zero margin to obstacles often results in collision in the real world. In scenarios where robots interact with people, certain additional conventional behaviors for navigation are expected which are not obvious from a traversability or safety standpoint, e.g., vehicles should drive on the right (or left) side of the road. For this particular example, semantics can be recognized as relating to the boundary between two adjacent concepts such as where the edge of the road and grass meet.  

In the case of path planning for vehicles, the optimal path is not always the collision-free path of shortest distance. This is especially true in the case of larger vehicles or in situations where robots and humans must interact. Additionally, there may be cultural or right-of-way considerations such as the desire to not trespass into private property where it can be avoided. Programmatically encoding each of these desiderata into the autonomy software for an UGV may be infeasible or imprecise. Learning desired navigation behaviors from human examples is a viable approach in these cases.


In our prior work, we have demonstrated approaches to reduce the time and effort required to assign labels to visual data and adapt visual classifiers to novel environments~\citep{Wigness16IROS,wigness2017unsupervised}. This paper describes the combination of these techniques with a method for learning robot control parameters through inverse optimal control (IOC). Our IOC solution uses the theory of maximum entropy to learn a reward function given environment features extracted from visual perception, and minimal ``optimal'' trajectory examples collected by a human driving the robot in the environment. We deploy the learned behavior on a robot and test at multiple navigation sites in a real-world environment. Figure~\ref{fig:overview} is an overview of the off-line reward function learning, and its use in on-line navigation. Lastly, we demonstrate the ability to learn new behaviors quickly in the field.

This article is an extension of our earlier conference paper~\citep{wigness2018robot}. In addition to a more thorough discussion of our approach, the main contributions of this extension include 1) additional field experiments for three learned behaviors, 2) the introduction of non-terrain feature maps for learning, and 3) an on-line human-in-the-loop capability.

\section{Related work}

Research in terrain traversability has been performed to provide details of an environment that mobile robots can use to make navigation decisions. \cite{papadakis2013terrain} provides a survey of work done in terrain traversability analysis. Most commonly, traversability analysis is treated as a binary classification problem, e.g., traversable versus non-traversable, for path planning~\citep{shneier2008learning,suger2015traversability,milella2015self}. Instances of multi-class terrain analysis have taken a supervised classification approach. Talukder et al. assume three terrain classes are known \textit{a priori} and train using Expectation Maximization to classify terrain for dynamic control of UGVs~\citep{talukder2002autonomous}. For multi-class problems, a continuous value relating to the cost of traversal for different terrains can also be assigned. These values have been assumed to be known~\citep{roncancio2014traversability} or can be learned~\citep{silver2008high}. Although it is likely that human operators demonstrating optimal trajectories are internally considering traversal feasibility for the platform, learning behaviors is more complex than a binary traversability problem. 

Self-supervised learning or learning from experience has been performed by maintaining traversal history and evaluating bumper collisions~\citep{shneier2008learning, Lookingbill07IJCV} for navigation tasks. This does not require any human intervention, but is only reliable in situations when collisions are minimal and present no risk to the robot. Imitation learning has been leveraged by collecting expert annotation of desirable traversal routes from aerial imagery~\citep{silver2008high}, and having humans teleoperate a mobile robot to collect desirable traversal examples~\citep{suger2015traversability}. The problem of learning from demonstrations has been addressed for situations requiring replication of behavior. Solutions based on inferring the agent's underlying reward structure have the advantage of being more succinct than the policy-based approaches and are therefore preferable as generalization becomes important~\citep{abbeel-icml-04,ziebart-AAAI-08,kitani-eccv-2012}. 

Most similar to our work is that of \cite{silver2008high}, who use maximum margin planning~\citep{ratliff-icml-06} to learn costs given desirable trajectory examples. However, this method performs poorly with imperfect behavior observations or when the feature space cannot perfectly describe the observed behavior. Our work addresses this limitation by replacing maximum margin planning with an approach based on the principle of maximum entropy~\citep{ziebart-AAAI-08}. Additionally, while \cite{silver2008high} uses overhead imagery which presents coarse-grained details of the terrain type, our work uses a visual classifier at ground level that can differentiate many types of terrain. Similarly, operation at ground level provides a limited view of the environment and unlike the case of aerial imagery, our example trajectories do not have long range knowledge of ``optimal'' paths in the environment. 

A potential limitation of the maximum entropy approach used in our work is the use of a linear reward function, which  may lack the representational power for some behaviors. \cite{levine-nips-11} developed GPIRL, which applies Gaussian Process Regression to learn nonlinear reward functions. Similarly, \cite{choi2013bayesian} proposed a Bayesian nonparametric approach to find a limited set of nonlinear rewards by learning a set of composites of logical conjunctions of atomic features. \cite{wulfmeier2016watch} describe an approach to learn cost maps from a large number of human driving demonstrations using maximum entropy deep inverse reinforcement learning with Fully Convolutional Neural Networks to represent the reward function. Although these approaches outperform the original maximum entropy framework~\citep{ziebart-AAAI-08}, they require a larger number of examples to learn a reward function. The nature of our application requires an approach that can adapt to situations involving smaller sets of training examples.

\section{Inverse optimal control methodology}

The approach we have developed for learning perception and control for robot navigation from human demonstration leverages many components from our prior work which is detailed in~\cite{Gregory16FSR,Wigness16IROS}. In this section, we focus on the key components specifically developed for this work. These components include a methodology for learning reward functions from human examples, and how to build occupancy feature maps from data taken from the robot's onboard sensors.

\subsection{Reward function learning}

In this work we generate motion models from examples and use them to plan paths for a robot that replicate the behavior demonstrated. Work in optimal control theory has produced algorithms that can learn motion models based on the observation of trajectories demonstrated by people (or other agents) as they move~\citep{kitani-eccv-2012,ziebart-iros-09}. These motion models capture the preferences of movers as they react to certain features in the environment. These preferences are encoded using a reward function, which describes how much a person favors taking a certain path over others. 

The problem of recovering this reward function can be solved using Inverse Optimal Control, which is also called Inverse Reinforcement Learning (IRL)~\citep{abbeel-icml-04}. In our work, we apply a specific solution to IRL based on the principle of maximum entropy that was originally proposed by \cite{ziebart-AAAI-08}. The advantage of this approach is that it resolves the ambiguity in choosing a distribution over decisions when finding a reward function. It selects the distribution that does not exhibit any additional preferences beyond matching feature expectations. In other words, it selects distributions based on what is known from the observations, without making any assumptions about what is not known. Moreover, the maximum entropy approach is well suited to the nature of our application, which focuses on the ability to quickly train for different behaviors in the field. By being no more committed to any particular path except the ones which have been demonstrated, it is possible to find reward functions without requiring a large number of examples.

The following sections provide a description of the elements of IRL that are more  relevant to our work. 

\subsubsection{Preliminaries}
The decisions made by a moving agent are modeled using a Markov Decision Process (MDP). An MDP is a tuple $\left( S,A,\left\lbrace P_{ss'}^{a}\right\rbrace,R,\gamma \right)$ where $S$ is a set of \textit{states}; $A$ is a set of \textit{actions}; $\left\lbrace P_{ss'}^{a}\right\rbrace$ is a set of \textit{transition probabilities}, where $P_{ss'}^{a}$ represents the probability of transitioning from state $s\in S$ to state $s'\in S$ after taking action $a\in A$; $R(s,a):S\times A\rightarrow \Re$ is the \textit{reward} function, which represents how much the agent is recompensed by taking an action $a$ when it is in a given state $s$ (though this work only considers reward functions that are independent of an action); and $\gamma \in [0,1)$ is a \textit{discount factor} that represents how much a reward in the future is worth compared to receiving the same reward now.

The behavior of an agent under this MDP is defined by a \textit{policy} $\pi : S \rightarrow A$, where the function $\pi (s)$ determines the action an agent should take in state $s$. Similarly, $\zeta$ is a \textit{path} taken by an agent under this MDP, and is a sequence of state and action tuples $\zeta = [(s_{1},a_{1}),(s_{2},a_{2}),\ldots]$. The probability of traversing a path is denoted $P(\zeta)$. Likewise, the reward function for taking a given path is 

\begin{equation} \label{eq:path_reward}
R(\zeta)=\sum_{(s_{i},a_{i})\in \zeta} R(s_{i}). 
\end{equation}

We also define a set of \textit{features} that represent the particular attributes of each state. Let $\phi: S\rightarrow \Re^D$, where $D$ indicates the dimensionality of the feature space. The feature vector representing state $s$ is $\phi (s)$. Additionally, the \textit{feature counts} of a path $\zeta$ is defined as 

\begin{equation} \label{eq:feature_counts}
 \phi_{\zeta}=\sum_{(s_{i},a_{i})\in \zeta} \phi (s_{i}).
\end{equation}

\subsubsection{Training a predictive model}
We now focus on the problem of learning a reward function (or \textit{training stage}) from a set of observations. The derivation presented in this section was reported by \cite{ziebart-AAAI-08} and is included here for a more complete description of our work. It is assumed that the observations were made from an agent behaving optimally, i.e. acting according to the optimal policy $\pi^*$. The reward function is defined as a linear combination of the features, i.e. $R=\theta^{T} \boldsymbol{\phi}(s)$. In this stage, we obtain the vector of weights $\theta$ of the optimal reward function. Per the maximum entropy distribution, paths that result in higher total reward should be exponentially more likely to be chosen. This concept is expressed as 

\begin{equation} \label{eq:dist_over_paths}
P(\zeta )= \frac{1}{Z(\theta)}\mathrm{exp}(R(\zeta ) ),
\end{equation}
where $Z(\theta)$ is the partition function. Assuming that transitioning randomness has little effect on behavior and the partition function is constant, a tractable approximation to Eq. (\ref{eq:dist_over_paths}) is given by: 

\begin{equation} \label{eq:apx_dist_over_paths}
P(\zeta \mid \theta) \approx \frac{\mathrm{exp}(\theta^{T}\boldsymbol{\phi}_{\zeta})}{Z(\theta)} \prod_{(s_{t},a_{t}),(s_{t+1},a_{t+1}) \in \zeta}P_{s_{t}s_{t+1}}^{a_t}
\end{equation}

This provides a stochastic policy, where an action is selected with a probability weighted by the sum of all probabilities of the paths taken that start with that action, as expressed by 

\begin{equation} \label{eq:stochastic_policy}
P(a \mid \theta) \propto \sum_{\zeta:a\in \zeta_{t=0}}P(\zeta \mid \theta).
\end{equation}

To find $\theta$ we maximize the log-likelihood of the distribution over paths (Eq. (\ref{eq:apx_dist_over_paths})). This is a convex optimization problem, where the optimal vector $\theta^*$ is given by 
\begin{equation} \label{optimal_theta}
\theta^{*}=\operatornamewithlimits{argmax}_{\theta}L(\theta)=\operatornamewithlimits{argmax}_{\theta}\sum_{i=1}^{N}\log P(\zeta_{i} \mid \theta),
\end{equation}
where $N$ is the number of training examples.

To solve this optimization problem \cite{ziebart-AAAI-08} express the gradient  as the difference between empirical feature counts and the learner's expected feature counts, which in turn can be expressed in terms of expected state visitation frequencies $D_s$:
\begin{equation} \label{eq:gradient}
\nabla L(\theta)=\hat{\phi}_{obs} - \sum_{i=1}^{N}P(\zeta \mid \theta)\phi_{\zeta} = \hat{\phi}_{obs}-\sum_{s\in S}D_{s}\phi(s)
\end{equation}

$D_s$ represents the probability of being in a given state $s$. It can be efficiently computed using dynamic programming. The transition probabilities can be approximated by the observed trajectories. \citeauthor{ziebart-AAAI-08} describe in detail the algorithm for the calculation of expected state frequencies.

\subsubsection{Trajectory generation}
After finding the reward function, the distribution over trajectories (Eq. (\ref{eq:apx_dist_over_paths})) is used to predict a trajectory and fed to the planner. We apply the methodology described in~\cite{ziebart-iros-09} to generate a path. However, since our planner requires only one goal, we don't reason about all possible destinations, though the smoothing procedure is still used to generate the destination prior distribution.

\subsection{Environment features} \label{sec:features}

Environment features are encoded as binary occupancy grid maps, where a grid cell in a feature map denotes the presence/absence of the feature it is modeling. Most of the feature types used in this work are modeled from sensor data, including obstacle features from LiDAR, and $m$ terrain classes identified in the environment via semantic segmentation. However, any feature that can be encoded as a binary occupancy grid could be used for learning. In Section~\ref{sec:landmine}, we discuss scenarios where \textit{a priori} information about the environment is encoded to model potentially dangerous traversal regions.

\begin{figure}
\centering
\includegraphics[width=\columnwidth]{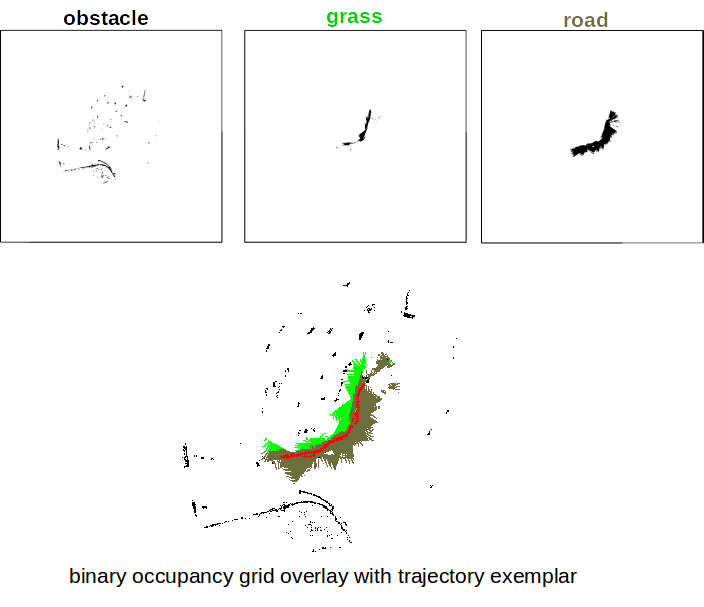}
\caption{(Top) Example binary occupancy grids for three feature types; obstacle collected by LiDAR and grass and road collected via semantic segmentation. (Bottom) An illustration of the obstacle grids overlaid one another with the corresponding demonstrated trajectory seen in red that was driven with respect to these environment features.}
\label{fig:feature grid_examples}
\end{figure}

Figure~\ref{fig:feature grid_examples} provides an example of three binary occupancy grid maps that encode the features \textit{obstacle}, \textit{grass}, and \textit{road}. The individual maps for each feature type can be seen in the top of the figure, where black denotes the presence of the feature and white denotes its absence. The color coded combination of these occupancy maps is provide in the bottom of the figure with a demonstrated trajectory overlaid in red. Specific details on how these feature grids are produced is provided in the remainder of this section.

Each binary occupancy grid feature map is output from a SLAM system based on the \emph{OmniMapper} described in~\cite{Trevor14ICRA}. Briefly, our SLAM system maintains a pose-graph along the robot's trajectory. Edges in this graph consist of measurements between consecutive robot poses in the form of odometry and 3D Generalized Iterative Closest Point (G-ICP)~\citep{Segal2009RSS}, in addition to loop-closure measurements through ICP when the robot revisits a previously explored location. The pose graph solution also incorporates measurements from GPS as in~\cite{Rogers14ACC}, allowing for long-term error correction in the absence of loop closures, and a global frame-of-reference to enable navigation to previously surveyed waypoints.  

Keyframes are collected along the robot's trajectory to store relevant sensor data for building binary occupancy feature grid maps. The obstacle feature map is produced from a Velodyne HDL-32E LiDAR, which is also used for G-ICP in consecutive keyframes and loop-closures. Keyframes are also created to establish the pose of the robot's cameras at each image taken to perform semantic segmentation, which is used to generate terrain feature maps. If the robot's trajectory is corrected through a loop closure or GPS measurement, keyframe poses are updated, resulting in newly rendered corrected binary occupancy feature maps. 

Semantic segmentation output at each camera keyframe is projected into the grid feature map by finding a bounding box of grid cells that project into the image. This is done by finding intersections between a flat ground plane and the extents of the image. Corner points of each candidate cell are projected into the image, and all pixels within the resulting quad are ``voted" into the occupancy grid based on their classification label. The label with the most votes sets that cell in the corresponding feature map; the cells in the terrain feature maps are entirely disjoint. This technique allows a cell to be updated by all images which observe it, and closer observations are naturally weighted higher due to the larger number of pixels in those images. Keyframes for laser data are limited to occur at no less than $50cm$ of distance traveled from the last keyframe to balance map quality with processing load. Visual keyframes are added at the frame rate of the visual semantic segmentation algorithm.

\begin{figure*}
\centering
\includegraphics[width=\linewidth]{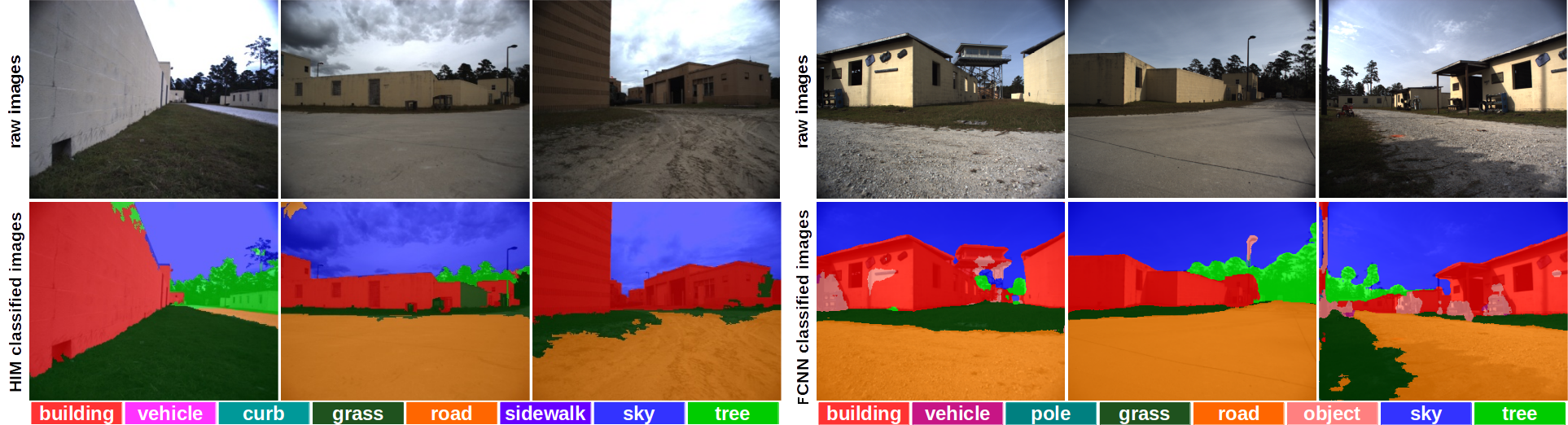}
\caption{Visual classification output by (Left) the HIM and (Right) the FCNN during on-line navigation. These frames are projected into the ground plane to generate binary occupancy grids for terrain feature types.}
\label{fig:classified_imgs}
\end{figure*}

Two different semantic segmentation algorithms are used throughout the experimental evaluation. Initial experiments (Sections~\ref{sec:roadgrass} and~\ref{sec:covert}) use the semantic segmentation output from a trained Hierarchical Inference Machine (HIM)~\citep{munoz13thesis}. The HIM trains a hierarchy of regressors to predict per pixel label output, and works well with a comparatively smaller set of training data compared to deep learners~\citep{simonyan2015very,szegedy2015going,krizhevsky2012imagenet,badrinarayanan2015segnet,long2015fully,zhou2014learning}. Efficient label collection techniques~\citep{Wigness16IROS,wigness2017unsupervised} are used to reduce human labeling intervention for classifier re-training and adaptation. The left portion of Figure~\ref{fig:classified_imgs} shows the label set and a selection of frames classified by the HIM in the real-world testing environment B.

The HIM is limited to processing speeds around $0.5 Hz$ for visual frames with resolution 344x275. Although this proves to be sufficient for real-time operation, deep learning techniques that perform inference on GPUs are able to provide significantly more visual perception information. For this reason, an implementation of a Fully Convolutional Neural Network (FCNN)~\citep{long2015fully} 
is trained using the Caffe framework~\citep{jia2014caffe}. This model is run on a Jetson TX2 (used in experiments from Sections~\ref{sec:landmine} and~\ref{sec:online}) to produce semantic segmentation inference at a rate of around $2 Hz$. The right portion Figure~\ref{fig:classified_imgs} shows the label set and frames classified by the FCNN in environment B. Qualitatively, the accuracy of both techniques is similar, yet the improved processing speed of the FCNN provides more visual information to project into terrain feature maps, giving denser information during training and on-line operation.

\begin{figure}
\centering
\includegraphics[width=\columnwidth]{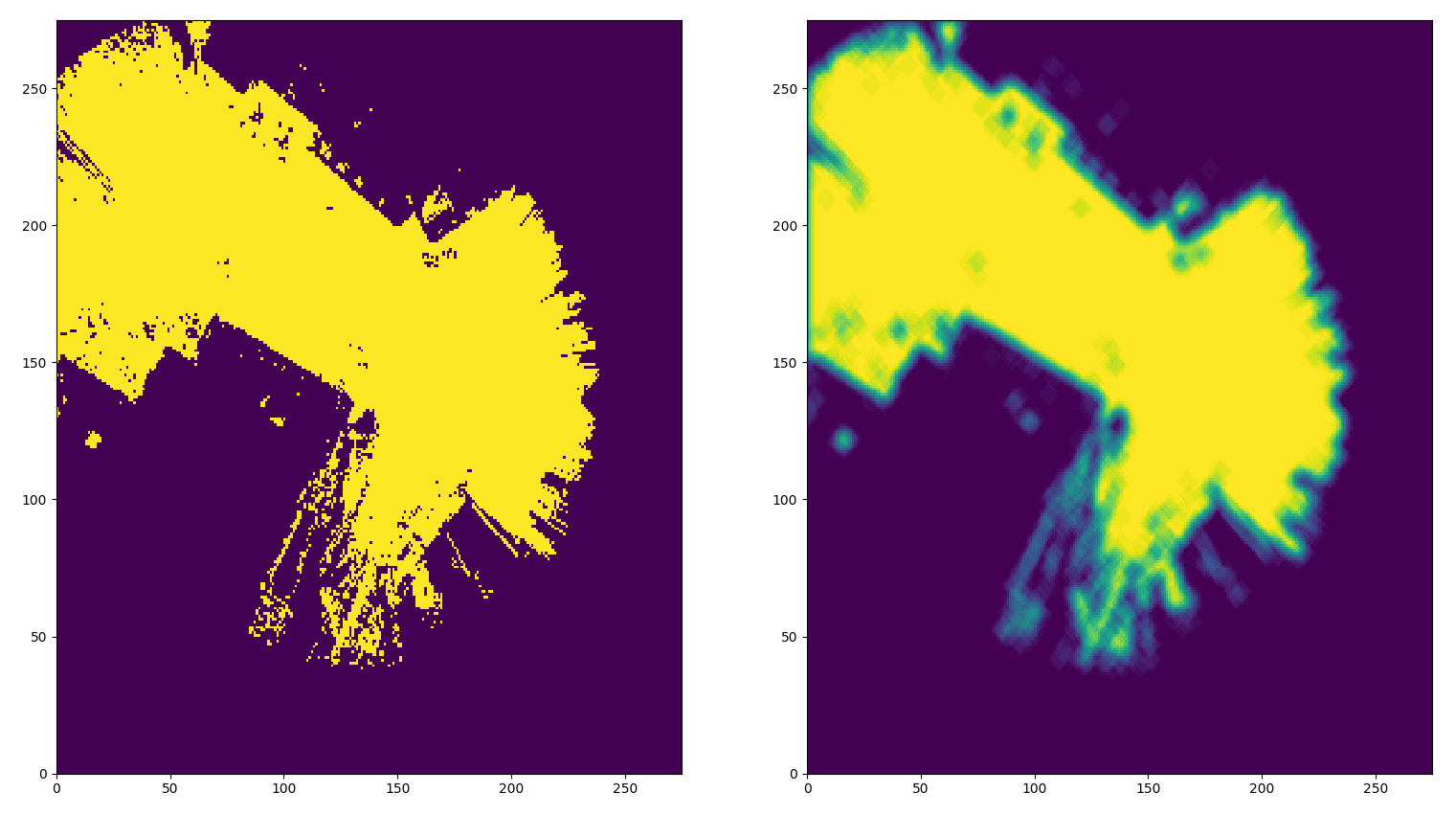}
\caption{A sample obstacle feature map (left) and its corresponding blurred feature map with a radius of 5 cells (right).}
\label{fig:sample_blurred_feat}
\end{figure}

In addition to the raw grid feature maps, several Gaussian blurred versions of these maps are used as additional features to encode the distance to the nearest cell of the classes. Figure~\ref{fig:sample_blurred_feat} shows an example obstacle map and a corresponding blurred feature map. The blurring effect extends the degree to which a cell represents the feature based on the values of cells within a specified radius. Cells closer to positive values, i.e., presence of the feature, in the grid map are weighted more heavily. Typically, for each feature we use a conventional grid map and at least one blurred map. Specific details regarding the feature set used in each experiment are provided in Section~\ref{sec:init}.

\section{Experiment setup}

\subsection{Outdoor environments}
Two urban environments are used to train and test our traversability cost learning with inverse optimal control. Environment A consists of a paved road intersection in an otherwise grassy wooded area, and is used as a training site to learn the first traversal behavior (discussed in Section~\ref{sec:roadgrass}). Environment B is a training facility meant to resemble a village in the developing world, with many small structures and gravel and dirt roads. Environment B is our primary testing environment, and is also used to collect data and train new behaviors in the field. 

\subsection{Human demonstrated trajectory collection}
\begin{figure}
\centering
\includegraphics[width=0.6\columnwidth]{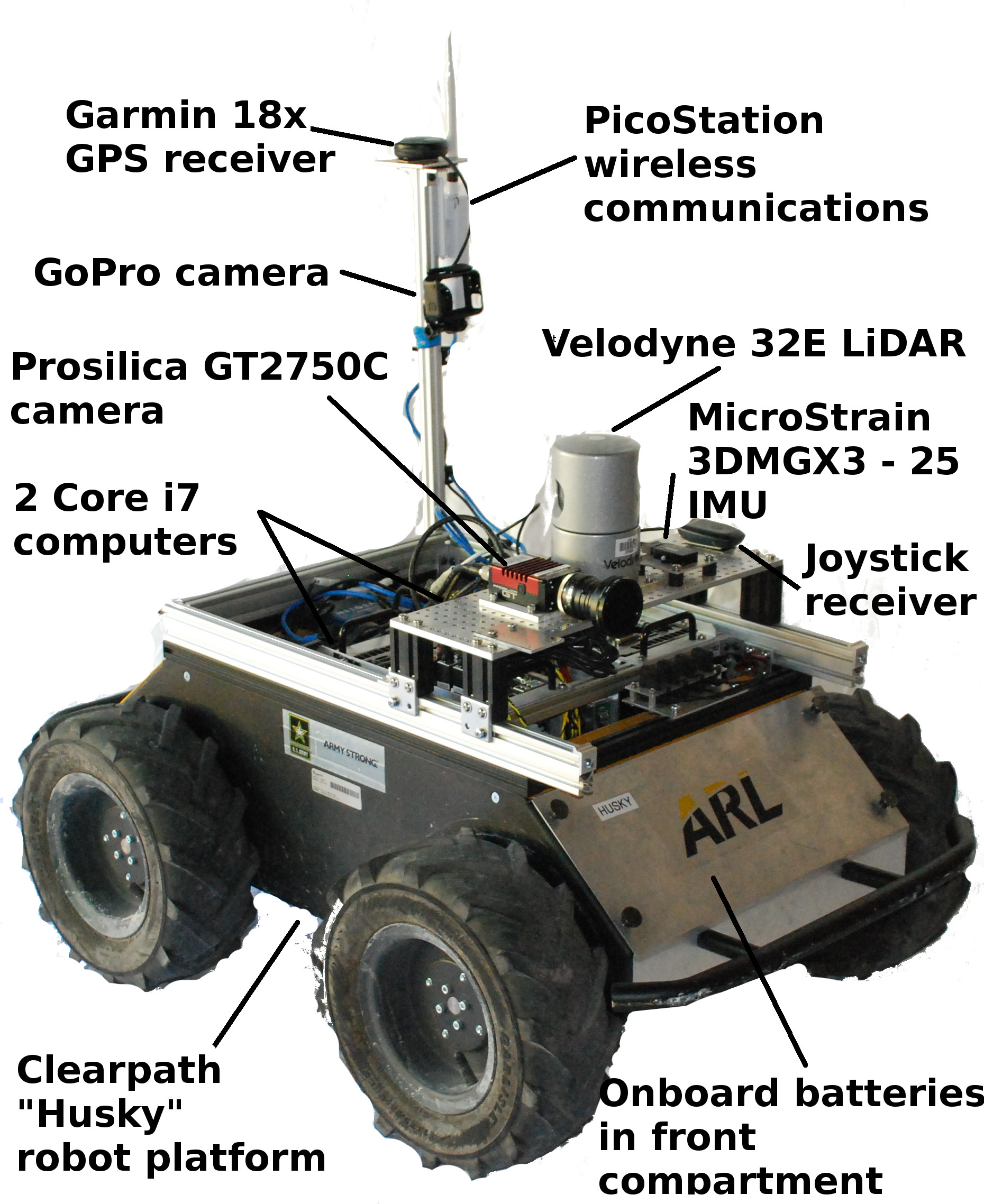}
\caption{The Husky robot with relevant sensor and components labeled.}
\label{fig:husky}
\end{figure}

Trajectory exemplars are collected using a Clearpath Husky, as seen in Figure~\ref{fig:husky}. This platform is also used to deploy the learned behavior during testing. The robot is teleoperated by a human from an initial starting point $i$ to a predefined goal destination $g$ with some human defined optimality. This optimality is defined based on the behavior that the operator would like the robot to learn. Section~\ref{sec:experiments} discusses and evaluates three learned behaviors, 1) traversal near a road edge, 2) covert traversal, and 3) traversal modification given intel relating to potentially dangerous areas.

Multiple demonstrated trajectories are collected for each behavior that is learned. Our training framework is capable of processing training exemplars with various feature map resolutions. This provides the added benefit of collecting demonstrations of varying trajectory length, and in different regions of the environment. Mapping and visual perception is run on-line during the trajectory demonstration to produce the binary feature maps for the relevant area of the environment.

\subsection{Training details} \label{sec:init}

The number of blurred maps and associated radii can be selected on a per-behavior basis. That is, some behaviors could be trained with many blurred feature maps with larger radii when feature information at larger distances is needed. The initial experiments for the edge of road traversal and covert traversal behaviors (discussed in Sections~\ref{sec:roadgrass} and~\ref{sec:covert}) were each performed with a feature set hand selected for the behavior. In addition to both behaviors learning a feature weight for obstacle, road, and grass features, a constant or bias feature is added to the feature weight vector. The blurred map radii used for each environment feature map for these behaviors is defined in Table~\ref{table:blur_info}.

However, in scenarios where a robot may need to rapidly switch between learned behaviors during on-line operation, it is ideal to have a standardized feature set, where each behavior is differentiated only by the values of its weight vector, $\theta$. Experiments discussed in Sections~\ref{sec:landmine} and~\ref{sec:online} all use learned reward functions based on a standardized feature set, which are summarized in the last two rows of Table~\ref{table:blur_info}.

When training without any prior initial weights, weights in $\theta$ are randomly initialized to values in the range of $[-5,5]$. The three experiments described in Section~\ref{sec:experiments} learn the behaviors from scratch with this random initialization. A previous set of weights can also be loaded to initialize $\theta$, which is the strategy used in our on-line experiments in Section~\ref{sec:online}. Each time the human interrupts the robot to provide a new trajectory demonstration, the weights currently being used for live operation are used to initialize $\theta$, and re-training is used to update these weights given the new demonstration.

\begin{table}
\footnotesize\sf\centering
\caption{Overview of the radii used to generate blurred feature maps for the environment features. A dash indicates that the environment feature was not used for the specified behavior.}
\label{table:blur_info}
\begin{tabular}{c|c|c|c|c}
\toprule
& \multicolumn{4}{c}{Blurred Map Radii} \\
Behavior & obstacle & road & grass & avoidance \\
\midrule
Edge of Road & 4 & 3, 6 & 3, 6, 9 & -- \\
Covert & 5, 10 & 5, 10 & 5, 10 & -- \\
ZOD Avoidance & 1, 2, 3, 4 & 1, 2, 3, 4 & 1, 2, 3, 4 & 1, 2, 3, 4 \\
On-line & 1, 2, 3, 4 & 1, 2, 3, 4 & 1, 2, 3, 4 & 1, 2, 3, 4 \\
\bottomrule
\end{tabular}
\end{table}

\subsection{Live operation experiment trials} \label{sec:trials}
In each navigation experiment, the robot must traverse from its initial location, $i$, to a waypoint goal, $g$, given GPS coordinate information. Selected $(i,g)$ pairs used in our experiments require on the range of 60 to 250 meters of traversal. For each experiment, a \textbf{ground truth (GT)} trajectory is collected according to $\pi^*$. Additionally, we run a \textbf{baseline} planner based on the search-based planning library (SBPL)~\citep{Cohen10ICRA} for comparison during each experiment. This planner generates a kinematically achievable global plan by searching combinations of motion primitives. This global plan minimizes traversal cost to the goal, and considers unknown grid cells to be of relatively high costs. This planner uses an unthresholded version of the obstacle feature map which additionally contains intermediate values for obstacles of varying laser ``opacity". This is accumulated in a negative log-odds grid based on the evidence of laser scans being interrupted by an obstacle in this cell against the evidence of which laser scans have passed through this cell. This planner does not make use of visual terrain features for planning. Finally, the \textbf{IOC} planner, when given a goal location, generates an output reward map. The reward map is then searched for an optimal reward trajectory which reaches the goal. This trajectory is passed to the navigation system of the robot which executes a kinematically feasible approximation of this trajectory.

Unless otherwise noted, four trials are run at each experiment site. Specifically, trial 1 captures trajectories of ground truth from $(i,g)$, IOC from $(g,i)$, and baseline from $(i,g)$. Trial 2 begins where trial 1 left off, i.e., from $g$, thus ground truth from $(g,i)$, IOC from $(i,g)$, and baseline from $(g,i)$. Trials 3 and 4 swap planner ordering to ground truth, baseline and IOC. By defining these four trials we are able to facilitate faster experimentation while collecting multiple trajectories for each technique in both directions of waypoint pairs. Further, this setup allows us to assume that coarse information about terrain and obstacles in the environment are known so the IOC and baseline planners have a larger field of view for navigation. By beginning each trial with a ground truth trajectory, this information can be populated in the map prior to the other planner execution. In future work this information could be obtained by teaming with an aerial vehicle that passes through the environment prior to the UGV, and providing an initial terrain classification from aerial imagery.

The trajectories for each trial are collected by recording the GPS measurements using the robot's onboard sensor as it moves between the two waypoints. Due to some GPS measurement drift\endnote[1]{We observed a GPS measurement bias of 2 or 3 meters at times due to signal blockage, resulting in minor abnormalities when the robot's position is plotted against a geo-referenced image.}, each trajectory starting point is re-aligned to the known surveyed waypoint start location. Trajectories for the third behavior experiment discussed in Section~\ref{sec:landmine} are collected using a Leico Viva TS16. The use of the total station allows us to eliminate the GPS drift issue for a more accurate trajectory comparison. The total station tracking prism is mounted on the middle of the right edge of the UGV main body, and tracked for each trajectory during an experiment trial.

\subsection{Evaluation metrics} 
Performance is measured by comparing a navigation trajectory, i.e., baseline or IOC, with a ground truth trajectory produced by human teleoperation. Trajectories are compared using the modified Hausdorff distance (MHD)~\citep{dubuisson1994modified,shao2010modified},
\begin{equation}
h(A,B) = \frac{1}{|A|} \sum_{a_i \in A} (\min_{b_j \in B} d(a_i,b_j)),
\end{equation}
where $B$ is the ground truth trajectory, $A$ is the trajectory to be compared, and function $d$ measures the distance between two points along these trajectories. This metric compares the geometry of trajectories, with lower MHD indicating higher similarity.

We present the mean and median MHD for the set of trials in each experiment, and the MHD for the best trial in each experiment. Each point along the trajectory is represented by its UTM coordinates, measuring distance in meters. The best mean MHD score is highlighted in bold during experimental analysis throughout this paper.

\section{Learned behaviors} \label{sec:experiments}


\subsection{Edge of road traversal} \label{sec:roadgrass}

For the first learned behavior, the optimal navigation policy ($\pi^*$) should maintain close proximity to grass but only traverse road terrain. This optimal behavior requires learning more than how to simply assign costs to distinct terrain types. Context of the environment with respect to the robot's task must also be considered. This behavior is chosen because it represents a traversal pattern that allows a mobile robot to adhere to common vehicular practice, i.e., driving on the road, but keeps the smaller platform away from potentially larger vehicles that may also be occupying the road.

\begin{figure*}
\centering
\includegraphics[width=.7\linewidth]{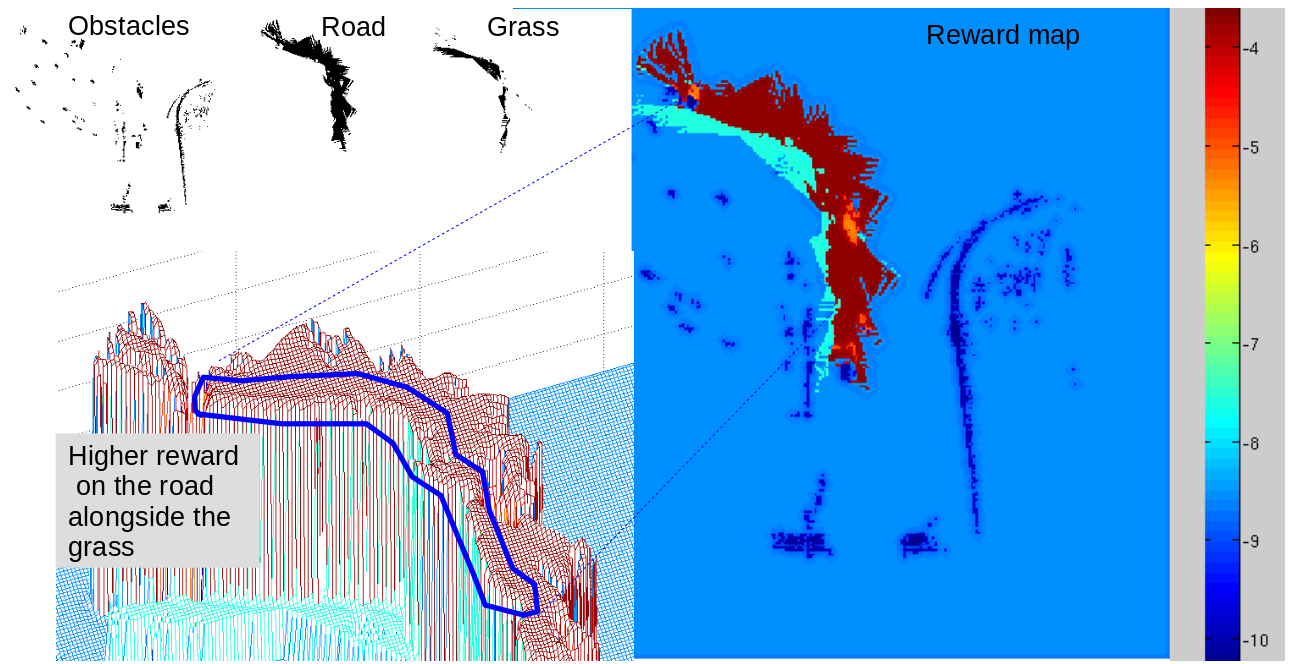}
\caption{Sample reward map obtained from demonstrated behavior. This figure illustrates the case of a robot that tends to drive on the road while staying close to the grass. As seen in the magnified insert (bottom left corner), this preference is indicated by a slight bump in the reward function.}
\label{fig:learned_reward_map}
\end{figure*}

\begin{figure}
\centering
\includegraphics[width=\linewidth]{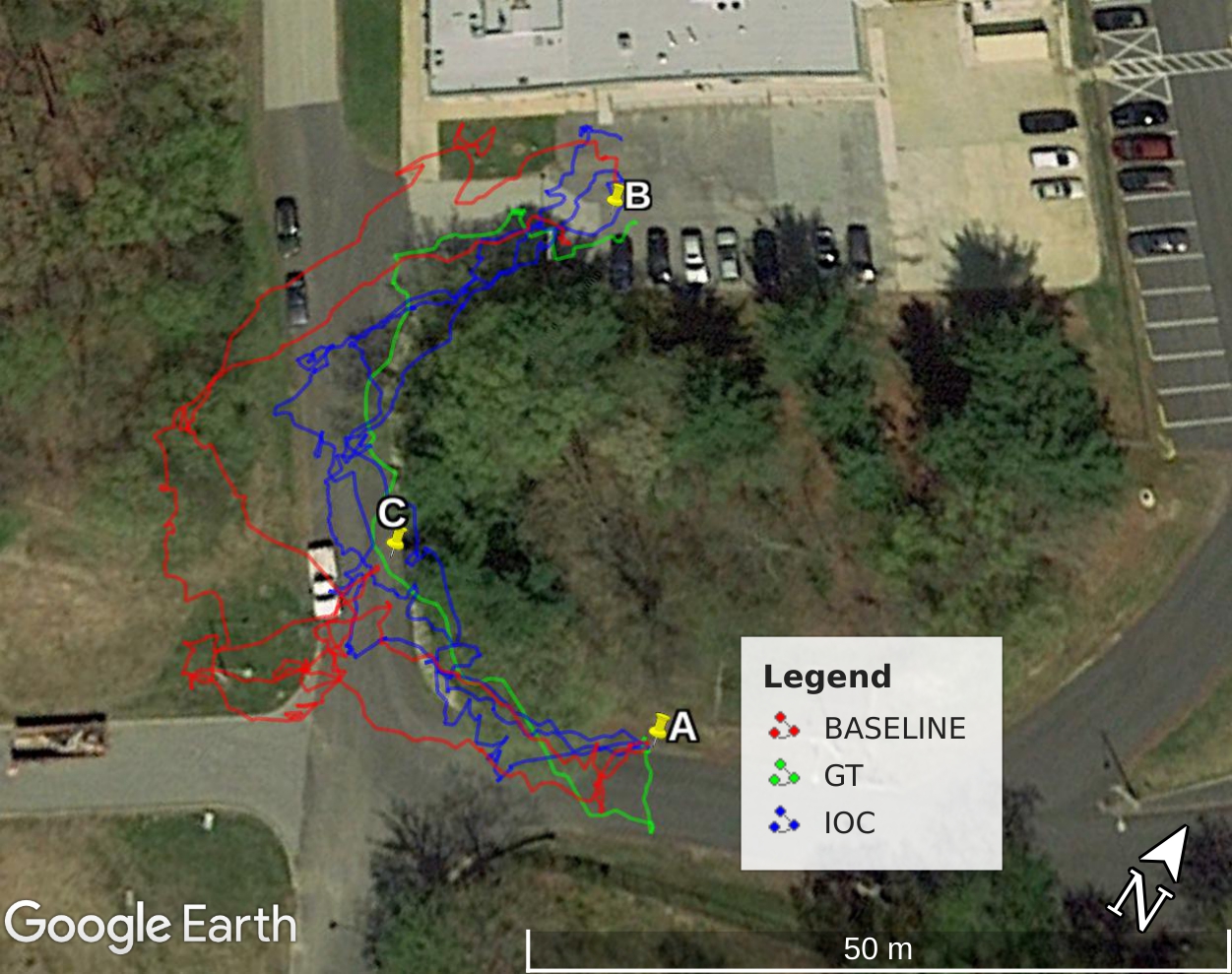}
\caption{Aerial view of environment A and the trajectories driven by the robot during two trials of the preliminary experimental evaluation of the learned behavior that focuses on maintaining close proximity to the edge of the road while navigating.}
\label{fig:aerial_envA2}
\end{figure}

\paragraph{\textbf{Environment A}} The training for this behavior is performed in environment A. Six exemplar trajectories are collected between the waypoint pair $(A,C)$ (referenced in Figure~\ref{fig:aerial_envA2}). Figure~\ref{fig:learned_reward_map} visually depicts the rewards learned for the features after training from trajectory examples and visual features collected in environment A. The individual feature maps are shown in the top left of the image, and an overlay of these feature maps are presented as a heat map on the right. Areas in the map that correspond to road have the highest reward seen in red, whereas obstacles have the lowest traversal reward seen in dark blue, and grass falls between these two features. The bottom left of the image is a magnified 3-D illustration of the map, which shows that not all areas of road terrain are represented by the same reward value. As demonstrated by the human trajectory exemplars, a higher reward can be seen on the edge of the road alongside the grass. This is visually indicated by the bump (outlined in blue) in the reward values on the map.

As a preliminary test, the learned costs for this behavior are deployed on the robot for on-line operation in environment A. Testing is performed in the same area as the training trajectories were demonstrated, but the navigation task is extended to be roughly twice the distance as that represented in the training trajectories. Seven testing trials are performed between the waypoint pair $(A,B)$ in environment A. Table~\ref{tab:envA-test} compares the MHD of the baseline and IOC planners at this testing sites with respect to the ground truth.

\begin{table}
\footnotesize\sf\centering
\caption{Comparison of the modified Hausdorff distance for the baseline and IOC planners with respect to ground truth trajectories near the original training site in environment A.}
\label{tab:envA-test}
\begin{tabular}{c|c|c|c|c|c|c}
\toprule
 & \multicolumn{3}{c|}{Baseline} & \multicolumn{3}{c}{IOC} \\
Test Site & Mean & Median & Best & Mean & Median & Best \\
\midrule
(A,B) & 4.114 & 3.488 & 1.447 & \textbf{3.036} & 2.827 & 1.376 \\
\bottomrule
\end{tabular}
\end{table} 

Across the seven trials our learned IOC planner produces a navigation trajectory that more closely resembles the optimal traversal behavior as defined by the collected GT trajectories. Although this initial test is performed in the same environment, test experiments were performed roughly two months after the collection of training data. This shows that the learned behavior is able to generalize to small changes in environment that naturally occur over time. Figure~\ref{fig:aerial_envA2} illustrates the trajectories from two of the testing trials to qualitatively show the closer trajectory aligned of the IOC planner to the ground truth than that of the baseline planner.

\begin{figure*}
\centering
\includegraphics[width=0.85\linewidth]{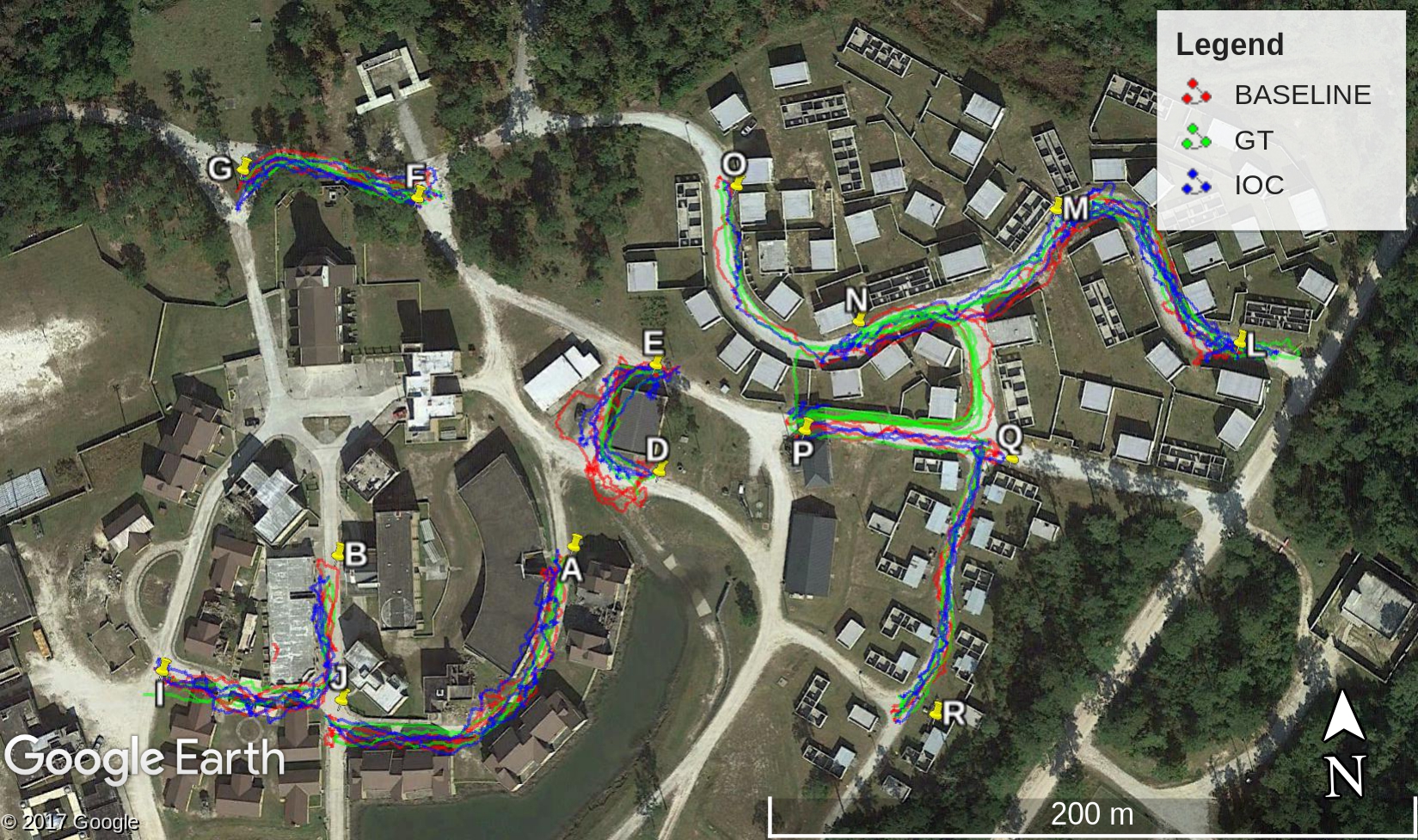}
\caption{Aerial view of environment waypoints and the trajectories driven by the robot during experimental evaluation of the learned behavior that focuses on maintaining close proximity to the edge of the road while navigating.}
\label{fig:aerial_envB}
\end{figure*}

\paragraph{\textbf{Environment B}}
To better show the generalization of the learned feature costs, we extensively test the learned behavior through a series of navigation experiments in environment B. Figure~\ref{fig:aerial_envB} is an aerial map of this environment, and depicts the test site waypoints and trajectories observed during evaluation. Each of the nine $(i,g)$ waypoint pairs are selected such that the shortest straight line trajectory between the waypoints does not follow $\pi^*$ used to collect training trajectories. The set of testing sites are diverse with respect to traversal distance, buildings, road width, and goal waypoint visibility. 

Table~\ref{table:envB-results} compares the MHD of the baseline and IOC planners at these nine testing sites with respect to the ground truth. The MHD results indicate that overall the IOC planner generates trajectories more similar to the ground truth than the baseline planner. This indicates that the IOC framework in fact learned $\pi^*$ well, and plans according to this policy even with training examples from a different domain.

\begin{figure*}
\centering
\includegraphics[width=\linewidth]{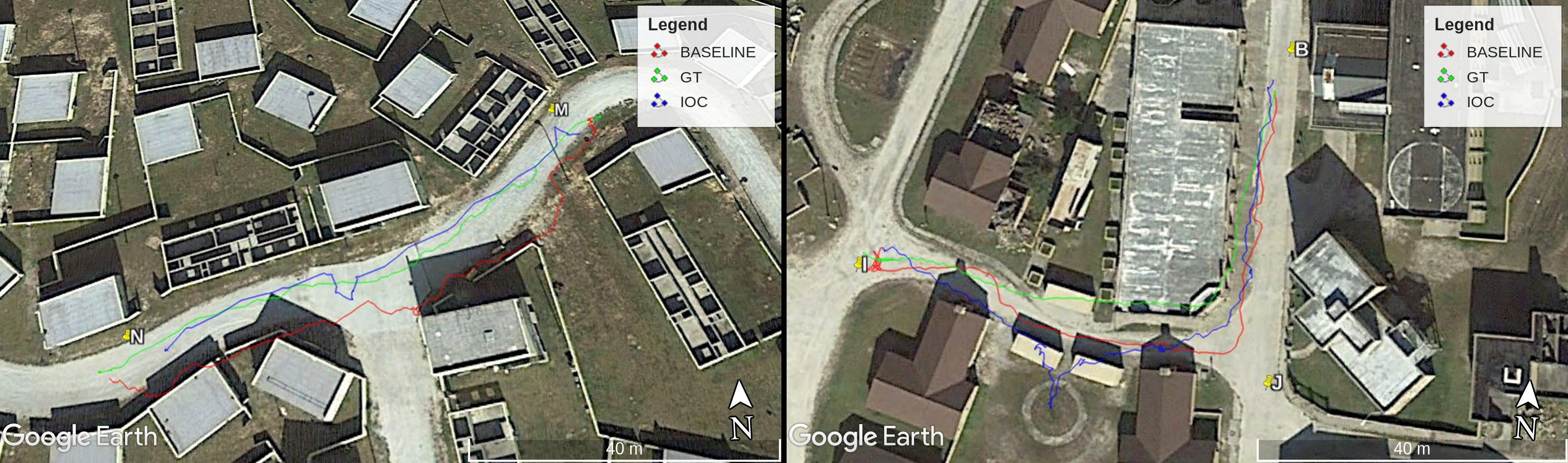}
\caption{Detailed views of two test runs. (Left) Test site $(M,N)$ which depicts a typical example of how the IOC planner closely mimics the GT trajectory while the baseline planner opts for paths far from obstacles. (Right) An example run from test site $(B,I)$ with poor performance from the IOC planner, where the grass feature was not found adjacent to the road, causing the robot to deviate off the expected path.}
\label{fig:qualitative}
\end{figure*}

A qualitative comparison of the trajectories output by the IOC and baseline planners allows us to visually see the learned behavior. The left image in Figure~\ref{fig:qualitative} show trajectories from one trial of the $(M,N)$ testing site. The trajectory overlays on the map show the close alignment of IOC and ground truth, whereas the baseline trajectory is much further away. 

However, Table~\ref{table:envB-results} also shows the baseline planner has a better median and best MHD than IOC at testing site $(B,I)$. The right image in Figure~\ref{fig:qualitative} shows an experiment trial from this site where the IOC trajectory actually deviates from the road, causing an increased MHD. We hypothesize that the poorer results at this testing site are caused by a less prevalent grass feature and narrower roadway than other testing sites.

\begin{table}
\footnotesize\sf\centering
\caption{Comparison of the modified Hausdorff distance for the baseline and IOC planners with respect to collected ground truth trajectories in environment B. *one trial and **two trial experiments.}
\label{table:envB-results}
\begin{tabular}{c|c|c|c|c|c|c}
\toprule
 & \multicolumn{3}{c|}{Baseline} & \multicolumn{3}{c}{IOC} \\
Test site & Mean & Median & Best & Mean & Median & Best \\
\midrule
(D,E) & 5.908 & 5.636 & 4.122 & \textbf{3.648} & 3.460 & 3.233 \\ \hline
(B,I) & \textbf{3.222} & 3.698 & 1.713 & 4.174 & 3.597 & 2.263 \\ \hline 
(A,J) & 4.242 & 3.786 & 3.530 & \textbf{3.626} & 3.836 & 1.025 \\ \hline
(F,G) & 2.641 & 2.413 & 1.871 & \textbf{1.738} & 1.844 & 1.018 \\ \hline
(Q,R) & 2.443 & 2.470 & 1.341 & \textbf{2.003} & 1.988 & 0.723 \\ \hline
(L,M) & 2.913 & 3.169 & 1.178 & \textbf{1.748} & 1.373 & 1.344 \\ \hline
(P,Q)** & 1.496 & 1.496 & 1.246 & \textbf{1.130} & 1.130 & 1.086 \\ \hline 
(L,O)* & -- & -- & 3.604 & -- & -- & \textbf{1.993} \\ \hline
(M,N)* & -- & -- & 4.682 & -- & -- & \textbf{2.164} \\ 
\bottomrule
\end{tabular} 
\end{table} 

We note that although in this image the IOC trajectory appears to pass through the buildings as it deviates from the road, this was not observed in the live experiments. Our UGV uses a Garmin GPS 18x single frequency (L1) receiver for global positioning. GPS signals are typically broadcast with $< 0.715m$ range error\endnote[2]{\url{www.gps.gov/systems/gps/performance/accuracy}}, but the GPS receiver solution can be corrupted by signal blockage (i.e. buildings, trees) and atmospheric conditions. In practice, we observed up to 2 or 3 meters of bias in some locations near buildings when the robot's position is plotted against a geo-referenced image. By subtracting this offset from the surveyed waypoint, this bias is removed from our evaluated trajectories. In the future, we intend to switch to a Hemisphere GNSS positioning system\endnote[3]{\url{https://hemispheregnss.com/Products/Products/Position/r330e284a2-gnss-receiver-760}} which incorporates the L2 signal, in addition to the other global positioning satellite constellations and multiple land-based corrections. 

\subsection{Covert traversal} \label{sec:covert}

One major benefit of the maximum entropy IOC learning approach is that reward functions can be learned with a relatively small number of training examples. This was demonstrated in the previous behavior with the use of only six demonstrated trajectories. This provides a way to quickly adapt a current behavior or learn a completely new behavior with minimal interruption in the robot's mission efforts. Our second experiment demonstrates the speed in which an alternative behavior can be learned and deployed with our IOC learning.

More specifically, this second experiment was performed end-to-end over the course of the last few hours of our field experimentation. This includes the collection of new training data, re-training of the reward function, and deployment of this behavior on the robot for live operation. The second learned behavior is described as ``covert" traversal. During this behavior, the robot should traverse in such a fashion that it is not openly acknowledged or visible. In this case, buildings, objects or other structures could be used as cover. We collect four new training demonstrations that keep the robot close to building edges and out of more visible, open areas such as roadways. Training data is collected from an area in environment B disjoint from any of the test waypoint locations. Using these new training examples, feature weights are re-learned and the robot is deployed to waypoint site $(P,Q)$ for testing.

\begin{figure}
\centering
\includegraphics[width=\columnwidth]{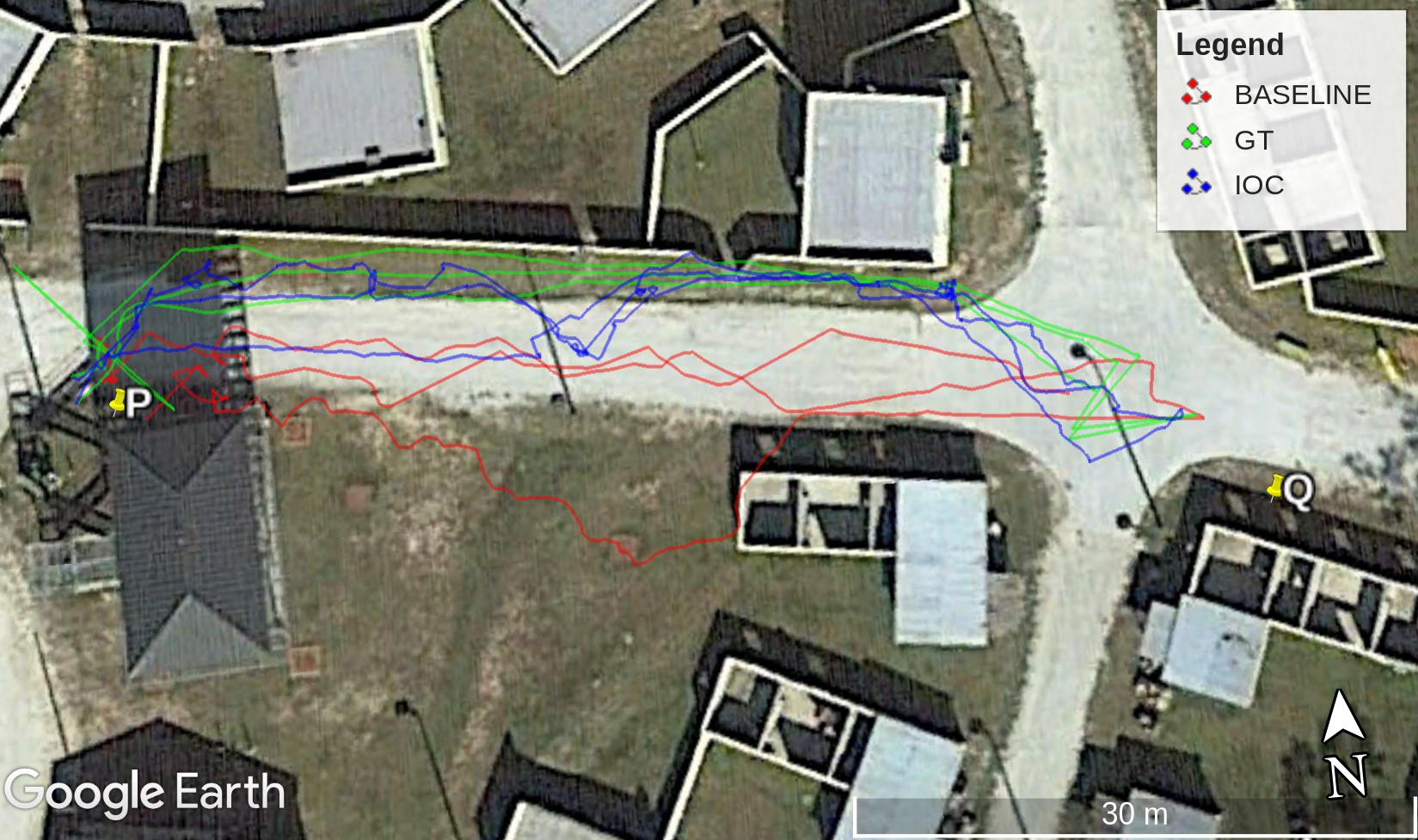}
\caption{Experiment trajectories$^1$ from P to Q using the new ``covert" training.}
\label{fig:run_p_q_covert}
\end{figure}

Table~\ref{table:newwights-results} shows the MHD results for the baseline and IOC planners for the covert traversal experiments. As in the previous experiments, ground truth is collected prior to running the IOC and baseline planners. Results show that the IOC planner performs more similarly to the ground truth than the baseline, suggesting the covert behavior has been learned well. Figure~\ref{fig:run_p_q_covert} also illustrates the trajectories of the trials for this experiment, showing qualitatively, the similarity between the ground truth and IOC.

\begin{table}
\footnotesize\sf\centering
\caption{Comparison of the modified Hausdorff distance for the baseline and IOC planners with respect to ground truth trajectories depicting covert traversal behavior.}
\label{table:newwights-results}
\begin{tabular}{c|c|c|c|c|c|c}
\toprule
 & \multicolumn{3}{c|}{Baseline} & \multicolumn{3}{c}{IOC} \\
Test Site & Mean & Median & Best & Mean & Median & Best \\
\midrule
(P,Q) & 5.201 & 4.457 & 3.422 & \textbf{1.415} & 1.362 & 1.160 \\
\bottomrule
\end{tabular}
\end{table} 

The MHD results also indicate that covert behavior is significantly different than a normal traversal behavior depicted by the baseline planner. In the first traversal behavior experiments, the IOC planner achieves only a 0.366 mean performance improvement over the baseline at waypoint site ($P,Q)$ (seen in Table~\ref{table:envB-results}). The performance gap jumps to 3.786 at the same site during the covert behavior experiment. This emphasizes the importance of learning behavior relevant to the current mission requirements or state of the dynamic environment, as normal behavior could deviate strongly from what is necessary to successfully complete tasks.

\subsection{Dangerous zone avoidance traversal} \label{sec:landmine}
Previous experiments focused on learning traversal behaviors related to environment features identifiable from LiDAR and camera sensor data. This perception information grounds the behavior learning to the context of physical objects and terrains in the environment. However, for certain applications, optimal task performance may require additional information beyond perception. For example, it may be desirable for a robot to stay within a specific communication radius. Conversely, there may be areas in an environment believed to present potential danger to a robot. Given known locations of communication nodes/dangerous zones, this information can be encoded as a binary occupancy grid indicating if grid cells are within the desired communication/dangerous range or not. 

\begin{figure*}
\centering
\includegraphics[width=.8\linewidth]{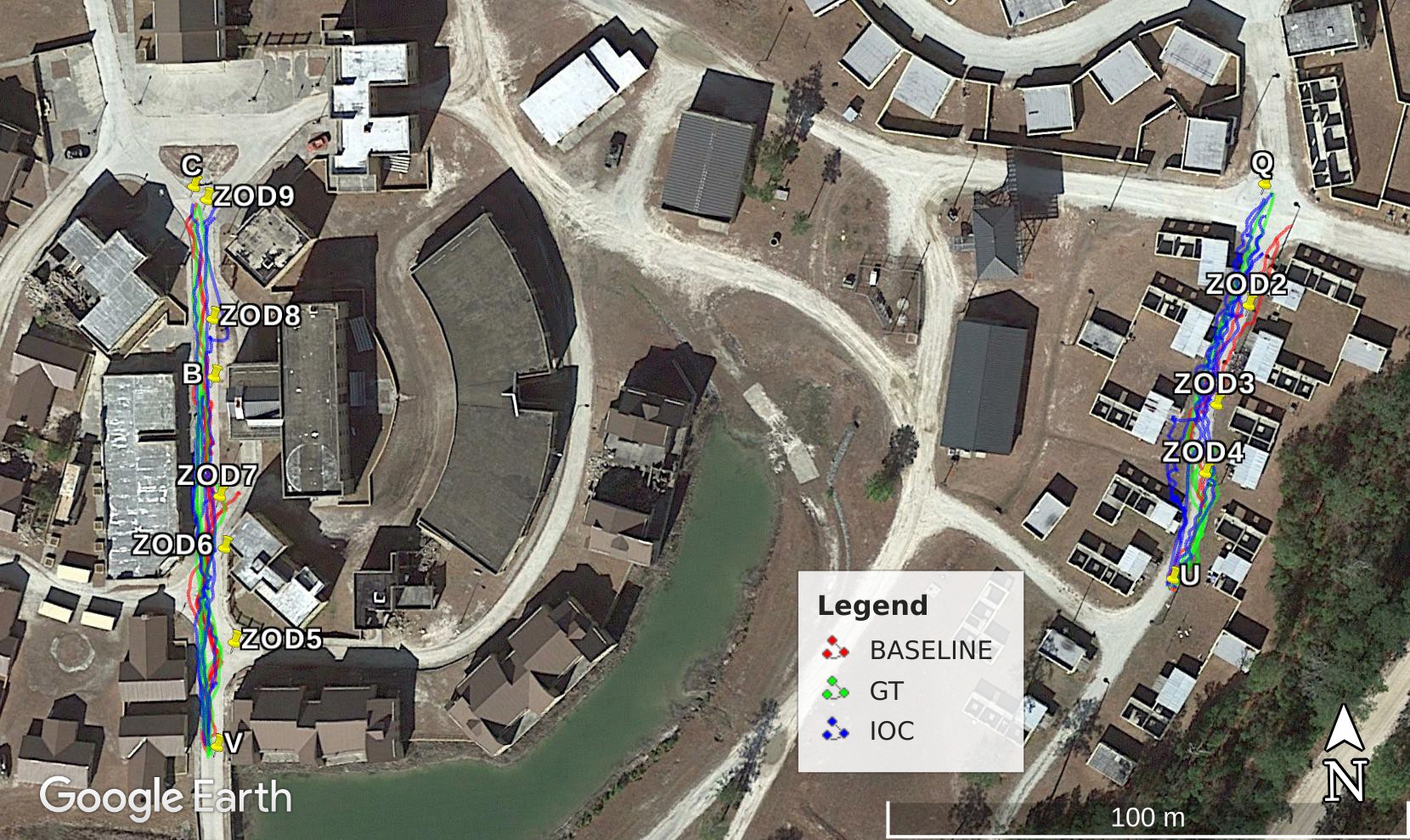}
\caption{Aerial view of environment waypoints and the trajectories driven by the robot during experimental evaluation of the learned behavior that focuses on avoiding traversal through zones of danger but otherwise maintaining close proximity to the edge of the road while navigating.}
\label{fig:landmine_map}
\end{figure*}

Our third behavior experiment focuses on a scenario that requires a traversal behavior that learns from perception derived feature maps in addition to features that are provided as \textit{a priori} intel. The optimal robot behavior for this task should maintain close proximity to the road edge (as in the experiment from Section~\ref{sec:roadgrass}), but should deviate when a potential zone of danger (ZOD) is encountered. Specifically, this scenario is designed to emulate a robot operating in an environment that may contain explosive devices. The \textit{a priori} information about the potential location of these devices and their expected blast radius is encoded as an \textit{avoidance} binary occupancy feature map. 

Training and testing of this behavior was performed in environment B. Figure~\ref{fig:landmine_map} shows an aerial view of the environment with all testing trajectories, testing waypoint sites, and zones of danger used in this experiment. Each ZOD marker in the map indicates the center of a circular zone of danger. Table~\ref{table:landmine_radius} indicates the specific radius in meters $(m)$ that defines each ZOD. The location and radius of ZODs are defined specifically to cover large portions of the road, including areas near road edges to test edge of road traversal deviation when a dangerous zone is encountered.

\begin{table}
\sf\centering
\caption{Radius associated with each zone of danger (ZOD).}
\label{table:landmine_radius}
\begin{tabular}{c|c}
\toprule
radius ($m$) & ZODs \\
\midrule
2  & 2, 3, 4 \\
2.5 & 5 \\
3 & 6, 7, 8, 9 \\
\bottomrule
\end{tabular}
\end{table} 

\begin{figure}
\centering
\includegraphics[width=\columnwidth]{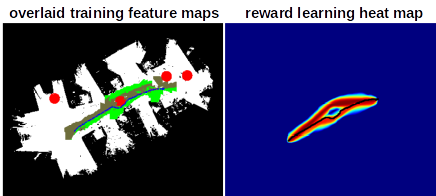}
\caption{Illustration of a training demonstration example and the corresponding reward map learned by our approach for the traversal behavior that avoids dangerous regions. (Left) An overlay of color coded feature maps, obstacle (black), road (gold), grass (green), and avoidance (red), with the exemplar trajectory shown in blue. (Right) The learned reward map for this exemplar is illustrated as a heat map, where hot colors indicate high reward. Notice that the location of the dangerous area (illustrated as red circles in the left image) is learned to be a low reward traversal region as indicated by the cooler colors in this area.}
\label{fig:landmine_reward_maps}
\end{figure}

12 training demonstrations were collected in the environment near waypoint Q, some of which were west of the testing waypoint site $(Q,U)$ (seen in Figure~\ref{fig:landmine_map}). Figure~\ref{fig:landmine_reward_maps} depicts one demonstrated trajectory exemplar and its corresponding reward map found during training. On the left is a color coded overlay of the feature maps used for this behavior including, obstacle (black), road (gold), grass (green), and avoidance (red). The demonstrated trajectory through this environment with these features is depicted by the blue line. The trajectory illustrates the desired behavior of traversing near the edge of the road until an avoidance region impacts this trajectory, forcing the traversal route to deviate around the ZOD into the grass before continuing along the edge of the road once the dangerous region has been passed.

The right image in Figure~\ref{fig:landmine_reward_maps} is the reward map found for this training example. The reward map is illustrated as a heat map, where red indicates the highest reward and blue the lowest. The demonstrated trajectory is also overlaid on the heat map to qualitatively show the rewards corresponding to this learned behavior. Notice that the reward map assigns a very low cost to the avoidance feature region, with higher rewards seen in all cells around this circular region. Further, although this particular demonstration deviated around the avoidance region by going ``below" this region, the reward function has learned that it is also reasonable to deviate over the ``top" of this region to mimic the optimal behavior.

This experiment represents a much more complex behavior than those demonstrated previously. For this scenario, there are really two underlying behaviors, i.e., traversing near the road edge and avoiding dangerous regions, that must be de-conflicted to perform the navigation task optimally. For this reason, we found that using more training exemplars than the previous behaviors provided better performance.

Evaluation of this behavior is performed at three waypoint sites in environment B. Table~\ref{table:landmine-mhd} compares the MHD scores for this behavior using our IOC planner and the baseline planner discussed previously. As seen the previous behaviors, the learned IOC reward function results in trajectories that closely resemble the ground truth trajectories that demonstrate the optimal behavior. This is seen in all cases (mean, median, and best of the trials), as IOC has a lower MHD than that of the baseline planner. We note that comparing the MHD of the IOC and baseline planners is not provided to show a ``superior" performance by the IOC for this behavior. Because the baseline planner is operating given only obstacle information, of course it will plan and have the robot traverse through the dangerous zones. Instead, the comparison is made to more fully indicate that the IOC planner is indeed learning the demonstrated behavior.

\begin{table} 
\footnotesize\sf\centering
\caption{Comparison of the modified Hausdorff distance for the baseline and IOC planners with respect to ground truth trajectories depicting traversal behavior that avoids dangerous regions. **two trial experiment}
\label{table:landmine-mhd}
\begin{tabular}{c|c|c|c|c|c|c}
\toprule
 & \multicolumn{3}{c|}{Baseline} & \multicolumn{3}{c}{IOC} \\
Test Site & Mean & Median & Best & Mean & Median & Best \\
\midrule
(Q,U) & 1.703 & 1.703 & 1.320 & \textbf{1.682} & 1.666 & 1.381 \\ 
(B,V)** & 1.047 & 1.047 & 0.904 & \textbf{0.775} & 0.775 & 0.720 \\ 
(C,V) & 1.263 & 1.251 & 1.127 & \textbf{1.100} & 1.204 & 0.677 \\ 
\bottomrule
\end{tabular}
\end{table} 

\begin{figure}
    \centering
    \includegraphics[width=\columnwidth]{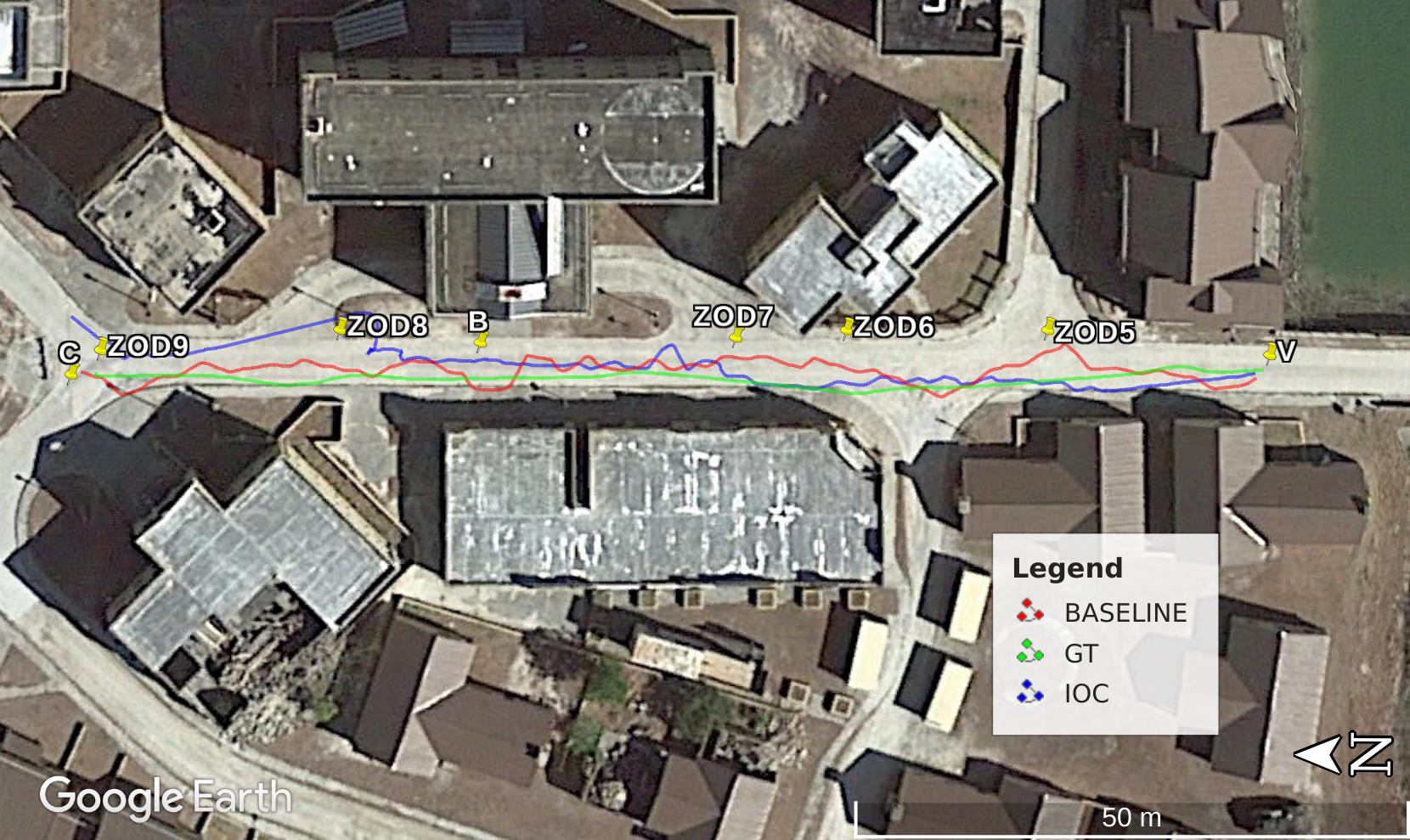}
    
    \includegraphics[width=\columnwidth]{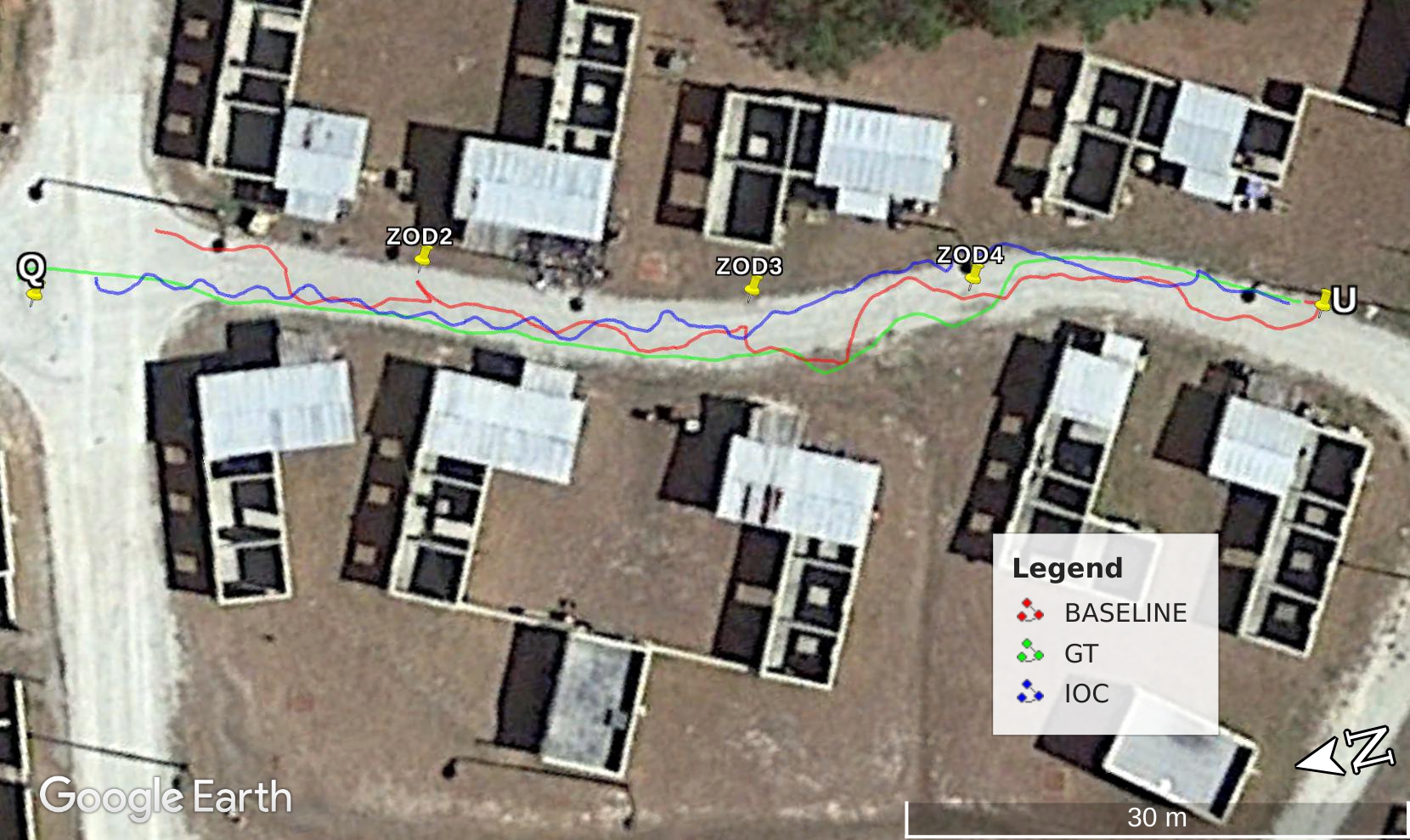}
    \caption{Qualitative illustrations of trajectories produced from human demonstration (GT), and the IOC and baseline planners during one trial of the dangerous zone avoidance traversal behavior experiments. (Top) Trial from test site $(C,V)$ with five ZODs to avoid. (Bottom) Trial from test site $(Q,U)$ with three ZODs to avoid.}
    \label{fig:landmine_trials}
\end{figure}

Figure~\ref{fig:landmine_trials} provides a qualitative comparison of trajectories from one trial at test site $(C,V)$ (top image), and one trial from test site $(Q,U)$ (bottom image). As expected, because the baseline planner has no knowledge of dangerous zones in the environment it traverses down the road, and enters the radius of ZOD5. However, the IOC planner has learned through the demonstrations that the radius of the dangerous zones should be avoided. While, the IOC trajectory does not deviate around the ZODs in strictly the same manner as the GT trajectory, the general behavior is still exemplified well. Part of the trajectory mismatch seen between IOC and GT in this trial run comes from the fact that the two trajectories are collected in different directions, i.e., GT from $(V,C)$ followed by IOC from $(C,V$) (as outlined in Section~\ref{sec:trials}. Because IOC begins at location $C$ it chooses to go around the ``top" side of ZOD8, but as IOC traverses closer to $V$ the trajectory more closely resembles the GT by deviating ``below" all other ZODs and also maintaining close proximity to the road edge.

Similar qualitative observations can be made about the trial run from $(Q,U)$ in the bottom image of Figure~\ref{fig:landmine_trials}. The IOC and GT trajectories at times take different deviations around a ZOD, but both trajectories do avoid the dangerous zone. This trial highlights that not only is the IOC planner avoiding the dangerous zones, but also maintains close proximity to the road edge in areas without ZODs. This is apparent when comparing the IOC trajectory to that of the baseline, which remains relatively centered in the road as it plans between the waypoints. Notice again that this causes the baseline trajectory to traverse extremely close to ZOD2 and ZOD4.

\section{On-line human demonstration and training} \label{sec:online}

In difficult or changing terrain, the ability to leverage corrective teleoperation provided by a human teammate could be used to enable on-line adaptation. Once a human teammate determines that the UGV is not exhibiting the desired behavior due to the presence of unexpected cues or changing environment conditions, they can pick up a joystick and demonstrate a desired trajectory. An on-line adaptive learning system can then record this trajectory segment together with the observed environmental features, and update its models appropriately while in the field and then resume operation.  

To enable this capability, we made several modifications to the existing learning from demonstration system. In the typical experimental setup, the left trigger on an XBox 360 joystick is depressed to override autonomous control with teleoperation. This type of teleoperation is used to maneuver to prepare for an experimental trial run or for platform safety. We have added a module which detects when this teleoperation trigger is depressed in addition to the right trigger. Depressing both triggers is used to indicate that a human teammate is providing a trajectory demonstration, and the trajectory is recorded until the triggers are released.

Once the human teammate has finished providing the new demonstrated training trajectory and autonomous control is restored to the UGV, the trajectory along with all feature maps (discussed in Section~\ref{sec:features}) are provided to an on-line training module. The on-line training module behaves similarly to the off-line procedure used in the previous experiments except that the weight vector $\theta$ is initialized with the current weights deployed on the UGV, and optimization is limited to 30 seconds to provide a quick on-line update to improve the platform navigation behavior. The on-line retraining procedure currently uses all previous training examples together with any new demonstrated trajectories and features. Once the optimization procedure converges or the allotted time elapses, the new weights are sent to update the control module. If the desired behavior is still not exhibited fully by the platform, the human teammate can interrupt UGV operation at any time to provide additional examples; in our experience a new behavior can be learned in as little as four examples.

\begin{figure*}
    \centering
    \begin{subfigure}[b]{\textwidth}
    \centering
        \includegraphics[width=0.45\textwidth]{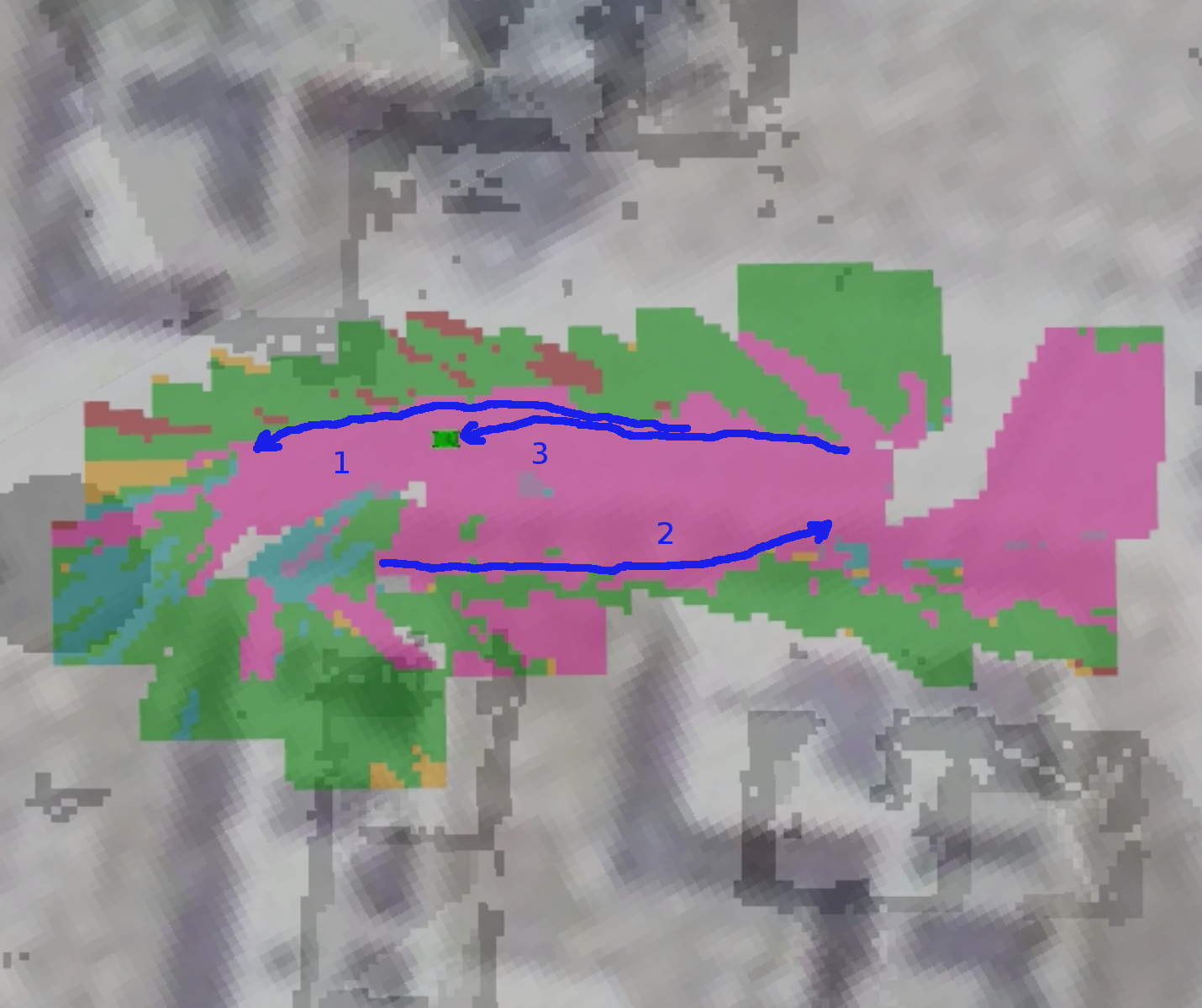}
        \includegraphics[width=0.43\textwidth]{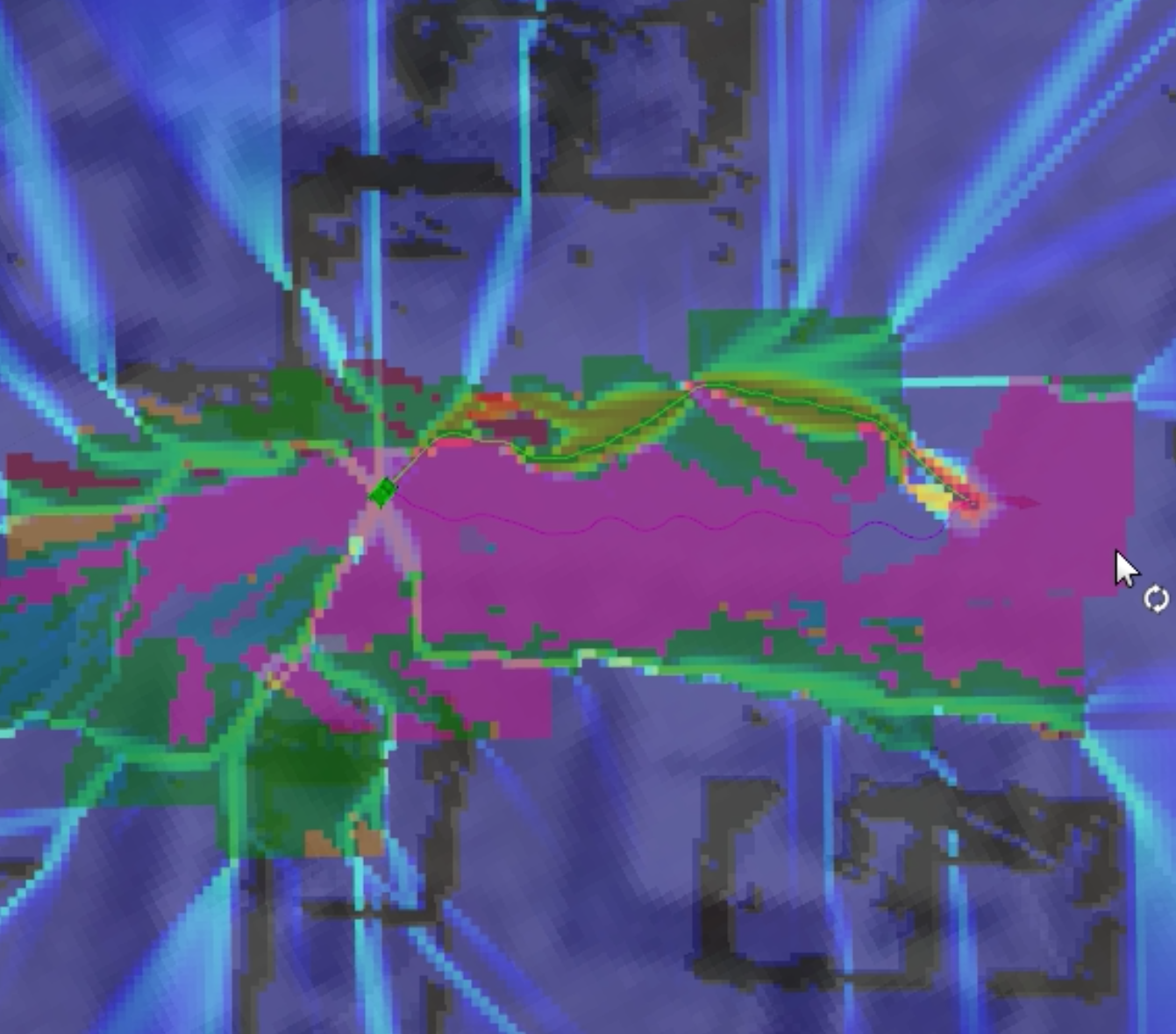}
        \caption{(Left) Three initial training trajectories are collected that demonstrate edge of road traversal. (Right) Testing results when running this behavior after these three demonstrations.}
        \label{fig:road_tr_phase1}
    \end{subfigure}
  
  \begin{subfigure}{\textwidth}
  \centering
    \includegraphics[width=0.45\textwidth]{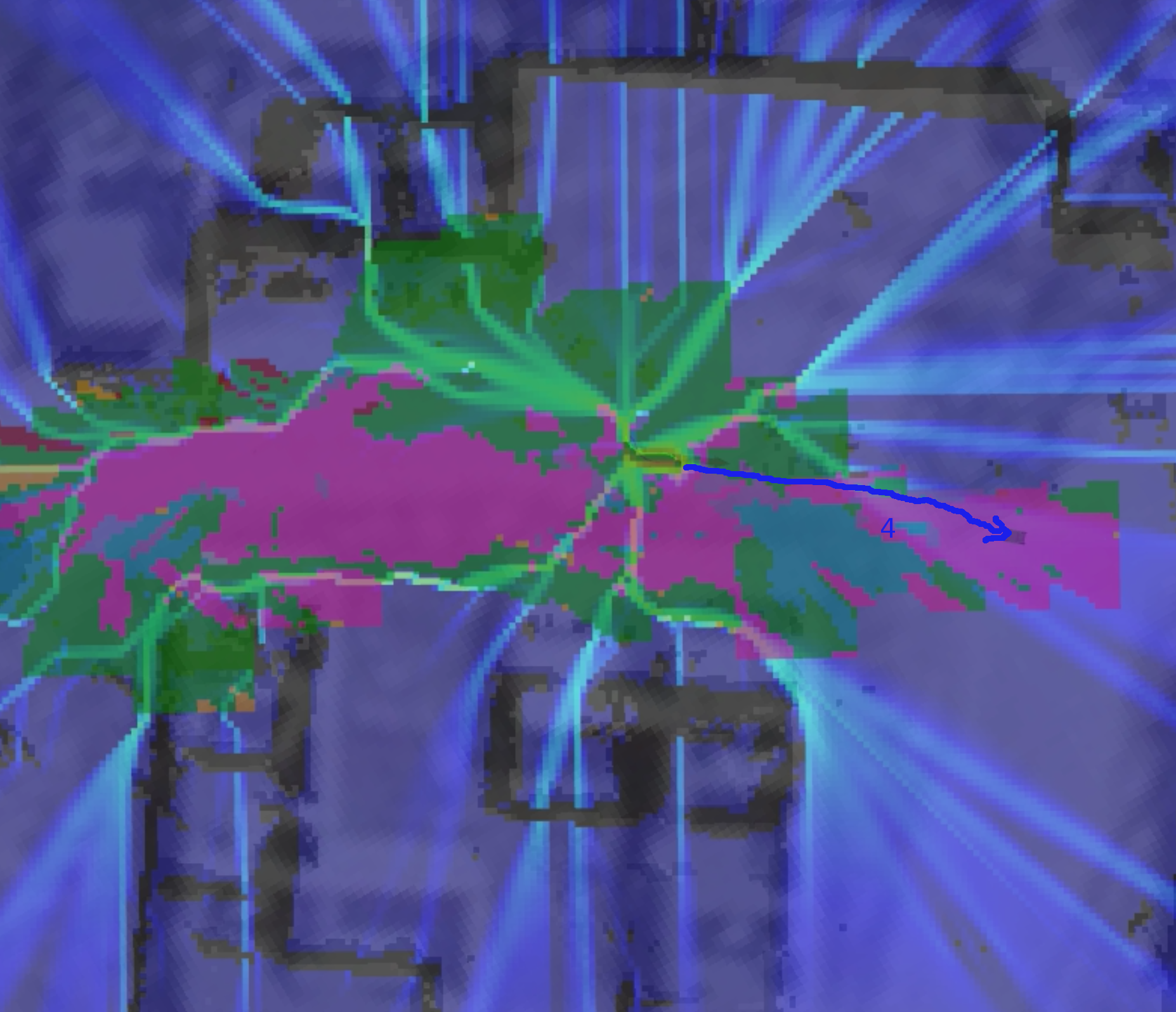}
    \includegraphics[width=0.425\textwidth]{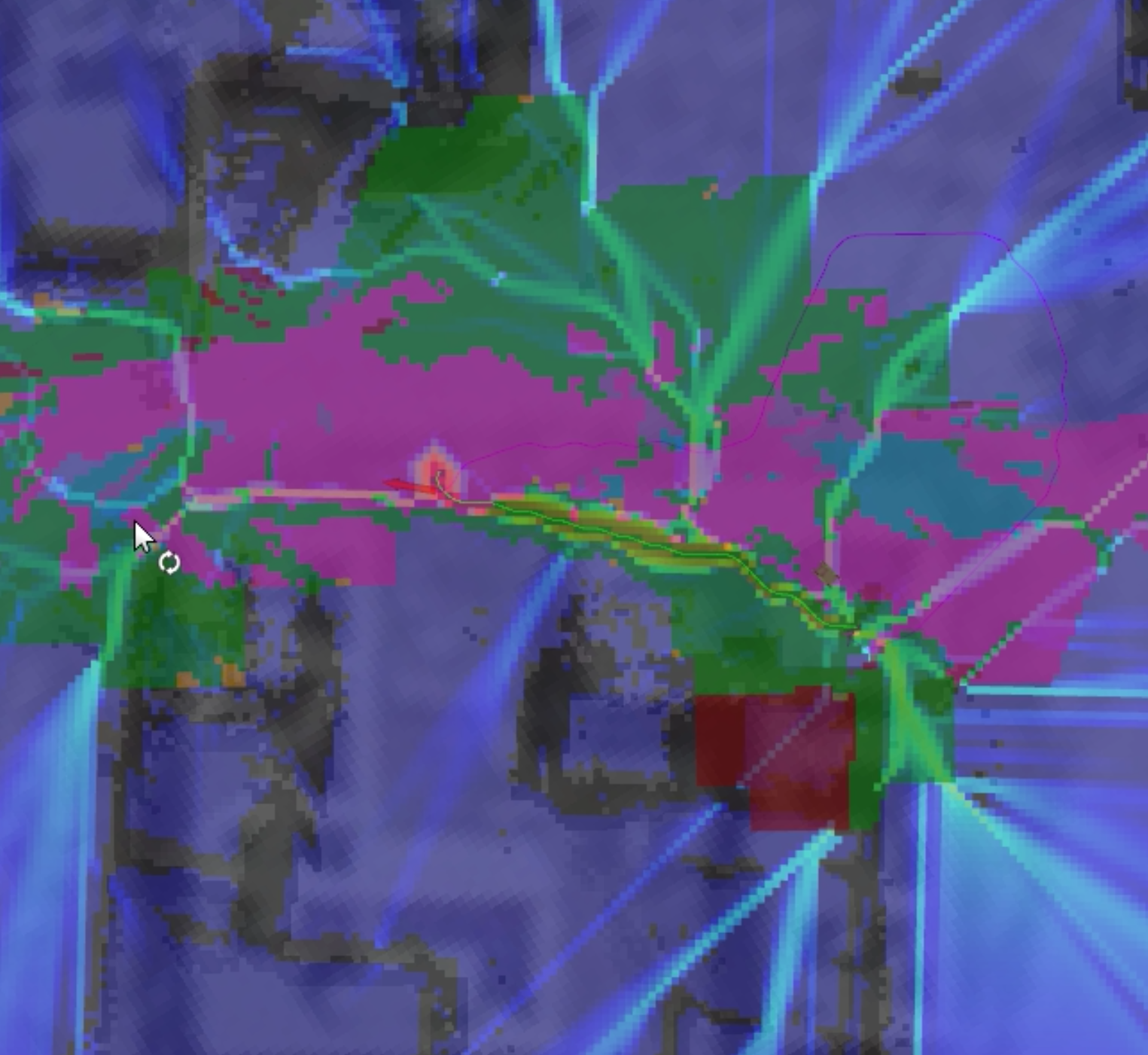}
    \caption{(Left) Collection of one initial training trajectory to improve the learned behavior of edge of road traversal. (Right) Testing results when running this behavior after the four demonstrations.}
        \label{fig:road_tr_phase2}
  \end{subfigure}
\caption{On-line demonstration interaction and testing operation of our IOC approach for the edge of road traversal behavior.}
    \label{fig:online-roadedge}
\end{figure*}

\begin{figure*}
\centering
    \begin{subfigure}[b]{\textwidth}
    \centering
        \includegraphics[width=0.475\textwidth]{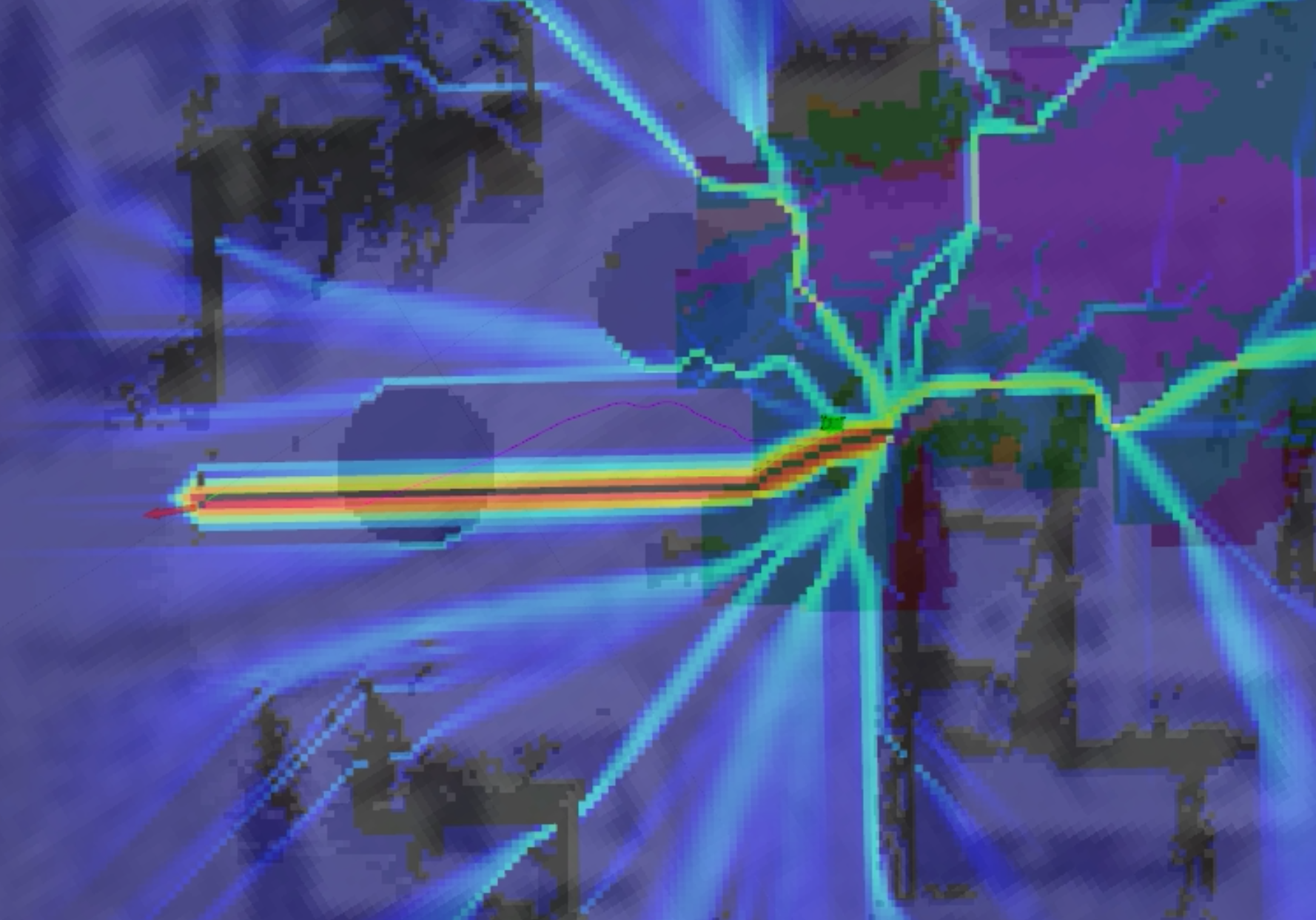}
        \includegraphics[width=0.415\textwidth]{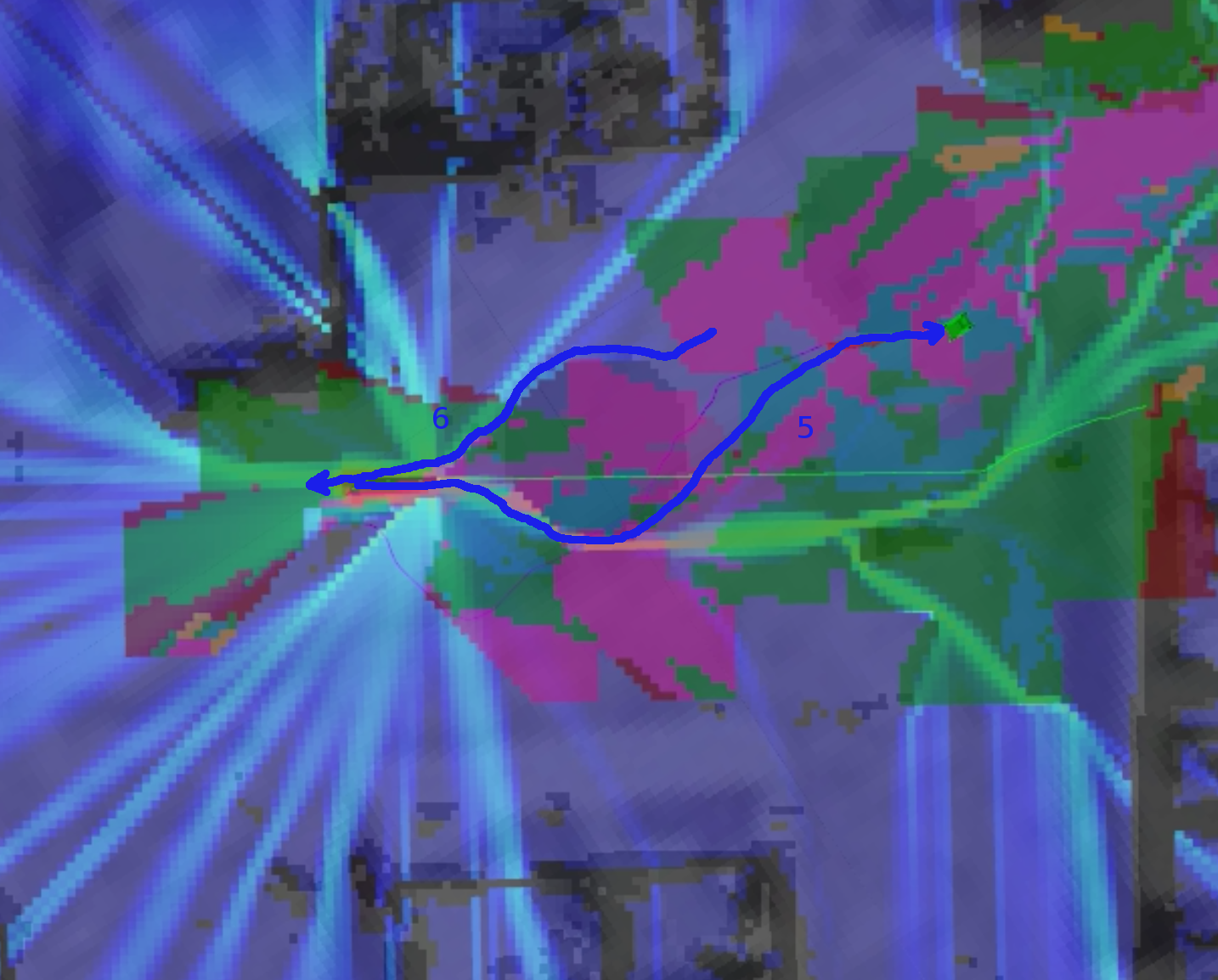}
        \caption{(Left) Testing results when running the behavior in an environment that now includes dangerous regions (shown as dark circles). Since the previous demonstrations were not based on this information, the planned path traverses straight through the dangerous region. (Right) Collection of two demonstrations that exhibit the behavior to traverse around these regions when present, otherwise maintaining close proximity to the road edge as demonstrated previously.}
        \label{fig:landmine_tr_phase1}
    \end{subfigure}
  
  \begin{subfigure}{\textwidth}
  \centering
    \includegraphics[width=0.475\textwidth]{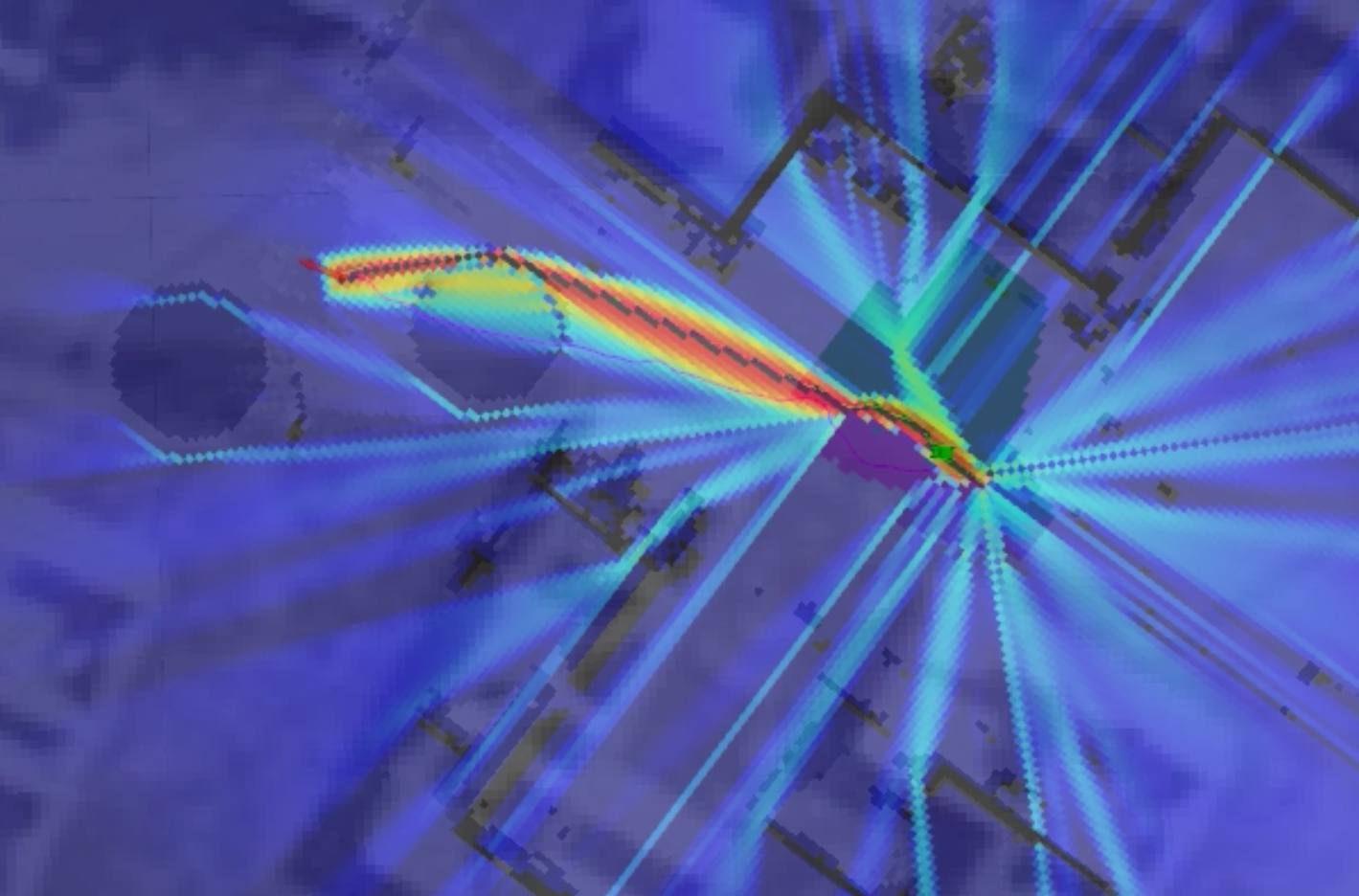}
    \includegraphics[width=0.425\textwidth]{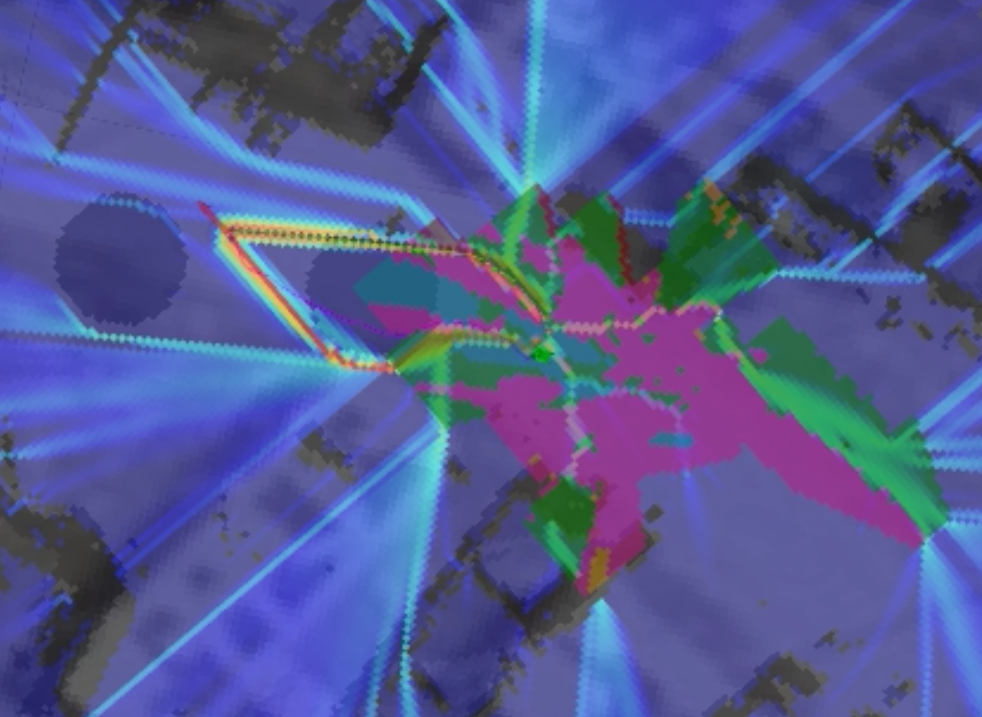}
    \caption{Two examples of testing results after the human demonstrations of traversing around the dangerous zones.}
        \label{fig:landmine_tr_phase2}
  \end{subfigure}
\caption{On-line demonstration interaction and testing operation when the robot is in a situation where it needs to change behaviors from edge of road traversal to now include dangerous region avoidance as well.}
    \label{fig:online-landmine}
\end{figure*}

An initial set of experiments were performed to determine the effectiveness of the on-line procedure. Figures~\ref{fig:online-roadedge} and~\ref{fig:online-landmine} illustrate several sequential steps of the on-line learning process for the scenario outlined in the rest of this section. These images show layers of information within the map used for navigation. The generated LiDAR obstacles are colored as grey/black regions. The visual perception terrain maps for road and grass are colored as pink and green, respectively. The reward heat map\endnote[4]{The illustration of reward heat maps show a visualization artifact that resembles lightning originating from the robot’s position. This effect is a result of using the Manhattan distance to calculate the blurred features and the destination prior distribution.} is overlaid in most of these images, indicating the path planned by the robot given its current set of feature weights. Dangerous zones are visualized as circular grey/black regions. Finally, demonstrated trajectories are drawn in blue. These visualizations are provided to show progression of learning that takes place with the on-line module. 

For this scenario, the UGV is assumed to start with no traversal behavior knowledge, i.e., no demonstrated training trajectories exist and thus, feature weights are randomly initialized (as outlined in Section~\ref{sec:init}). To begin, the human teammate determines that the UGV should behave in such a way that its traversal pattern follows the edge of the road (just as in the experiments from Section~\ref{sec:roadgrass}). As the robot has no understanding of this behavior the human teammate must provide some demonstration.

The human teammate depresses the trigger buttons, and provides three demonstrations of this behavior, which are labeled as $\left(1,2,3\right)$ in the left image of Figure~\ref{fig:road_tr_phase1}. The weight function is then automatically re-learned with these new examples onboard the UGV. The human teammate then asks the UGV to autonomously traverse to a specified location, but it exhibited unexpected behavior by driving into grass regions. This can be seen in the right image of Figure~\ref{fig:road_tr_phase1}, where the hot part of the heat map is completely ``above" the road terrain (shown in pink) and instead overlays the grass regions. To improve this behavior, a fourth training example was provided by the human teammate, as shown in the left of Figure~\ref{fig:road_tr_phase2}. The weights were updated onboard the UGV, and in a subsequent autonomous command the desired behavior of following the edge of the road was exhibited correctly, as seen in the right of Figure~\ref{fig:road_tr_phase2}, where the hot part of the heat map resides along the edge of the road terrain.

We then assume within this scenario that the environment and information about the safety of the environment changes. From the previous demonstrations, the robot is currently operating under the traversal behavior that maintains close proximity to the road edge, but now enters a section of the environment filled with areas that are dangerous to traverse. Absent any relevant training examples for this situation, the robot will plan autonomous trajectories that pass directly through these dangerous regions. This can be seen in the left of Figure~\ref{fig:landmine_tr_phase1}, where the dangerous zone is the dark circular region that the planned path crosses. The intel on these environment changes is provided to the robot as an additional feature, i.e., the avoidance feature map, and the human teammate seizes control of the UGV to provide two training examples, labeled as $\left(5,6\right)$ in the right image of Figure~\ref{fig:landmine_tr_phase1}, that exemplify how the robot should update its behavior in the presence of this new feature. 

These new training examples were incorporated into the learned model, and the weights were updated to reflect the cost of driving in the dangerous zones. Given this new training, the robot is assigned goals to test the updated behavior. Figure~\ref{fig:landmine_tr_phase2} shows two planned trajectories, and corresponding reward heat maps where the robot plans around these dangerous zones. The performance of avoiding these dangerous areas allows the human teammate to carry out other tasks relevant to the mission, while the robot autonomously and safely maneuvers to carry out its other higher level tasks.

This initial on-line experiment demonstrates the potential of our inverse optimal control methodology. With only a few demonstrated trajectories and limited training time, our system was able to learn feature weights that allowed the UGV to mimic the desired traversal behavior. This allows a robot to be deployed and updated in the field with minimal human interactivity. This is ideal for environments or missions that have the possibility to change rapidly during operation.

\section{Conclusions}

Several concerns need to be addressed when training UGVs to work alongside human teammates in unstructured environments, e.g., HA/DR or military operation. First, prior training may not be available or inadequate due to the chaotic nature of a scenario. This inadequacy can come in the form of visual appearance cues being altered so terrain cannot be evaluated correctly, as well as traversal cost changes i.e. dirt becomes mud which is harder to traverse. Secondly, the roles and activities envisioned by robot programmers might not encompass all of those which are needed in a dynamic and fast-evolving unstructured environment. An adaptable system will be able to contribute to the operation in more ways than a fixed one. By minimizing the effort needed to retrain or train a new behavior, a system will also reduce demands on personnel who are busy with other tasks.

This paper described an approach to building such a system which learns from human demonstration to acquire new skills with limited supervision in the field. The trained system was extensively evaluated in a real-world environment demonstrating three learned behaviors: 1) driving along the edge of the road, 2) covert traversal, and 3) avoidance of dangerous zones. The approach was shown to learn these behaviors with the use of feature maps built from visual perception input and \textit{a priori} intel. Further, the system was modified to allow human interruption to provide additional trajectory demonstrations and re-training updates during on-line operation. Currently, the types of skills learned are limited to navigational behaviors. However, with additional effort this approach should be applicable to other task domains. Future work will be to develop an end-to-end system that directly couples the visual classifier and IOC reward function without intermediate human-guided semantics. In this system, unsupervised clustering will replace visual classifier training, requiring only supervision in the form of example driving trajectories.

\theendnotes


\begin{thebibliography}{35}
\providecommand{\natexlab}[1]{#1}
\providecommand{\url}[1]{\texttt{#1}}
\providecommand{\urlprefix}{URL }
\expandafter\ifx\csname urlstyle\endcsname\relax
  \providecommand{\doi}[1]{DOI:\discretionary{}{}{}#1}\else
  \providecommand{\doi}{DOI:\discretionary{}{}{}\begingroup
  \urlstyle{rm}\Url}\fi

\bibitem[{Abbeel and Ng(2004)}]{abbeel-icml-04}
Abbeel P and Ng A (2004) Apprenticeship learning via inverse reinforcement
  learning.
\newblock In: \emph{Proceedings of the Int. Conf. on Machine Learning}. pp.
  1--8.

\bibitem[{Badrinarayanan et~al.(2017)Badrinarayanan, Kendall and
  Cipolla}]{badrinarayanan2015segnet}
Badrinarayanan V, Kendall A and Cipolla R (2017) {SegNet}: A deep convolutional
  encoder-decoder architecture for image segmentation.
\newblock \emph{Transactions on Pattern Analysis and Machine Intelligence} .

\bibitem[{Choi and Kim(2013)}]{choi2013bayesian}
Choi J and Kim KE (2013) Bayesian nonparametric feature construction for
  inverse reinforcement learning.
\newblock In: \emph{Proceedings of the Int. Joint Conference on Artificial
  Intelligence}. pp. 1287--1293.

\bibitem[{Cohen et~al.(2010)Cohen, Chitta and Likhachev}]{Cohen10ICRA}
Cohen BJ, Chitta S and Likhachev M (2010) Search-based planning for
  manipulation with motion primitives.
\newblock In: \emph{Proceedings of the Int. Conf. on Robotics and Automation}.
  IEEE, pp. 2902--2908.

\bibitem[{Dubuisson and Jain(1994)}]{dubuisson1994modified}
Dubuisson MP and Jain AK (1994) A modified {Hausdorff} distance for object
  matching.
\newblock In: \emph{Proceedings of the Int. Conf. on Pattern Recognition},
  volume~1. IEEE, pp. 566--568.

\bibitem[{Gregory et~al.(2016)Gregory, Fink, Stump, Twigg, Rogers, Baran, Fung
  and Young}]{Gregory16FSR}
Gregory J, Fink J, Stump E, Twigg J, Rogers J, Baran D, Fung N and Young S
  (2016) Application of multi-robot systems to disaster-relief scenarios with
  limited communication.
\newblock In: \emph{Proceedings of Field and Service Robotics}. Springer, pp.
  639--653.

\bibitem[{Jia et~al.(2014)Jia, Shelhamer, Donahue, Karayev, Long, Girshick,
  Guadarrama and Darrell}]{jia2014caffe}
Jia Y, Shelhamer E, Donahue J, Karayev S, Long J, Girshick R, Guadarrama S and
  Darrell T (2014) Caffe: Convolutional architecture for fast feature
  embedding.
\newblock \emph{arXiv preprint arXiv:1408.5093} .

\bibitem[{Kitani et~al.(2012)Kitani, Ziebart, Bagnell and
  Hebert}]{kitani-eccv-2012}
Kitani K, Ziebart B, Bagnell D and Hebert M (2012) Activity forecasting.
\newblock In: \emph{Proceedings of the European Conf. on Computer Vision}. pp.
  201--214.

\bibitem[{Krizhevsky et~al.(2012)Krizhevsky, Sutskever and
  Hinton}]{krizhevsky2012imagenet}
Krizhevsky A, Sutskever I and Hinton GE (2012) Imagenet classification with
  deep convolutional neural networks.
\newblock In: \emph{Proceedings of Advances in neural information processing
  systems}. pp. 1097--1105.

\bibitem[{Levine et~al.(2011)Levine, Popovic and Koltun}]{levine-nips-11}
Levine S, Popovic Z and Koltun V (2011) Nonlinear inverse reinforcement
  learning with gaussian processes.
\newblock \emph{Proceedings of Advances in Neural Information Processing
  Systems} : 19--27.

\bibitem[{Long et~al.(2015)Long, Shelhamer and Darrell}]{long2015fully}
Long J, Shelhamer E and Darrell T (2015) Fully convolutional networks for
  semantic segmentation.
\newblock In: \emph{Proceedings of the Conf. on Computer Vision and Pattern
  Recognition}. IEEE, pp. 3431--3440.

\bibitem[{Lookingbill et~al.(2007)Lookingbill, Rogers, Lieb, Curry and
  Thrun}]{Lookingbill07IJCV}
Lookingbill A, Rogers J, Lieb D, Curry J and Thrun S (2007) Reverse optical
  flow for self-supervised adaptive autonomous robot navigation.
\newblock \emph{Int. Journal of Computer Vision} 74(3): 287.

\bibitem[{Milella et~al.(2015)Milella, Reina and Underwood}]{milella2015self}
Milella A, Reina G and Underwood J (2015) A self-learning framework for
  statistical ground classification using radar and monocular vision.
\newblock \emph{Journal of Field Robotics} 32(1): 20--41.

\bibitem[{Munoz(2013)}]{munoz13thesis}
Munoz D (2013) \emph{Inference Machines: Parsing Scenes via Iterated
  Predictions}.
\newblock PhD Thesis, The Robotics Institute, Carnegie Mellon University.

\bibitem[{Murphy(2014)}]{murphy2014disaster}
Murphy RR (2014) \emph{Disaster robotics}.
\newblock MIT press.

\bibitem[{Papadakis(2013)}]{papadakis2013terrain}
Papadakis P (2013) Terrain traversability analysis methods for unmanned ground
  vehicles: A survey.
\newblock \emph{Engineering Applications of Artificial Intelligence} 26(4):
  1373--1385.

\bibitem[{Ratliff et~al.(2006)Ratliff, Bagnell and Zinkevich}]{ratliff-icml-06}
Ratliff N, Bagnell J and Zinkevich M (2006) Maximum margin planning.
\newblock In: \emph{Proceedings of the Int. Conf. on Machine Learning}. pp.
  729--736.

\bibitem[{Rogers et~al.(2014)Rogers, Fink and Stump}]{Rogers14ACC}
Rogers JG, Fink JR and Stump EA (2014) Mapping with a ground robot in {GPS}
  denied and degraded environments.
\newblock In: \emph{Proceedings of the American Control Conference}. IEEE, pp.
  1880--1885.

\bibitem[{Roncancio et~al.(2014)Roncancio, Becker, Broggi and
  Cattani}]{roncancio2014traversability}
Roncancio H, Becker M, Broggi A and Cattani S (2014) Traversability analysis
  using terrain mapping and online-trained terrain type classifier.
\newblock In: \emph{Proceedings of the Intelligent Vehicles Symposium}. IEEE,
  pp. 1239--1244.

\bibitem[{Segal et~al.(2009)Segal, Haehnel and Thrun}]{Segal2009RSS}
Segal A, Haehnel D and Thrun S (2009) Generalized-{ICP}.
\newblock In: \emph{Robotics: Science and Systems}, volume~2. p. 435.

\bibitem[{Shao et~al.(2010)Shao, Cai and Gu}]{shao2010modified}
Shao F, Cai S and Gu J (2010) A modified {Hausdorff} distance based algorithm
  for 2-dimensional spatial trajectory matching.
\newblock In: \emph{Proceedings of the Int. Conf. on Computer Science and
  Education}. IEEE, pp. 166--172.

\bibitem[{Shneier et~al.(2008)Shneier, Chang, Hong, Shackleford, Bostelman and
  Albus}]{shneier2008learning}
Shneier M, Chang T, Hong T, Shackleford W, Bostelman R and Albus JS (2008)
  Learning traversability models for autonomous mobile vehicles.
\newblock \emph{Autonomous Robots} 24(1): 69--86.

\bibitem[{Silver et~al.(2008)Silver, Bagnell and Stentz}]{silver2008high}
Silver D, Bagnell J and Stentz A (2008) High performance outdoor navigation
  from overhead data using imitation learning.
\newblock In: \emph{Robotics: Science and Systems}.

\bibitem[{Simonyan and Zisserman(2015)}]{simonyan2015very}
Simonyan K and Zisserman A (2015) Very deep convolutional networks for
  large-scale image recognition.
\newblock In: \emph{Proceedings of the Int. Conf. on Learning Representations}.

\bibitem[{Suger et~al.(2015)Suger, Steder and
  Burgard}]{suger2015traversability}
Suger B, Steder B and Burgard W (2015) Traversability analysis for mobile
  robots in outdoor environments: A semi-supervised learning approach based on
  {3D-LiDAR} data.
\newblock In: \emph{Proceedings of the Int. Conf. on Robotics and Automation}.
  IEEE, pp. 3941--3946.

\bibitem[{Szegedy et~al.(2015)Szegedy, Liu, Jia, Sermanet, Reed, Anguelov,
  Erhan, Vanhoucke and Rabinovich}]{szegedy2015going}
Szegedy C, Liu W, Jia Y, Sermanet P, Reed S, Anguelov D, Erhan D, Vanhoucke V
  and Rabinovich A (2015) Going deeper with convolutions.
\newblock In: \emph{Proceedings of the Conf. on Computer Vision and Pattern
  Recognition}. IEEE, pp. 1--9.

\bibitem[{Talukder et~al.(2002)Talukder, Manduchi, Castano, Owens, Matthies,
  Castano and Hogg}]{talukder2002autonomous}
Talukder A, Manduchi R, Castano R, Owens K, Matthies L, Castano A and Hogg R
  (2002) Autonomous terrain characterisation and modelling for dynamic control
  of unmanned vehicles.
\newblock In: \emph{Proceedings of the Int. Conf. on Intelligent Robots and
  Systems}, volume~1. IEEE, pp. 708--713.

\bibitem[{Trevor et~al.(2014)Trevor, Rogers and Christensen}]{Trevor14ICRA}
Trevor AJ, Rogers JG and Christensen HI (2014) Omnimapper: A modular multimodal
  mapping framework.
\newblock In: \emph{Proceedings of the Int. Conf. on Robotics and Automation}.
  IEEE, pp. 1983--1990.

\bibitem[{Wigness et~al.(2018)Wigness, Rogers and
  Navarro-Serment}]{wigness2018robot}
Wigness M, Rogers JG and Navarro-Serment LE (2018) Robot navigation from human
  demonstration: Learning control behaviors.
\newblock In: \emph{Proceedings of the International Conference on Robotics and
  Automation}. IEEE, pp. 1150--1157.

\bibitem[{Wigness et~al.(2016)Wigness, Rogers, Navarro-Serment, Suppe and
  Draper}]{Wigness16IROS}
Wigness M, Rogers JG, Navarro-Serment LE, Suppe A and Draper BA (2016) Reducing
  adaptation latency for multi-concept visual perception in outdoor
  environments.
\newblock In: \emph{Proceedings of the Int. Conf. on Intelligent Robots and
  Systems}. IEEE, pp. 2784--2791.

\bibitem[{Wigness and Rogers~III(2017)}]{wigness2017unsupervised}
Wigness M and Rogers~III JG (2017) Unsupervised semantic scene labeling for
  streaming data.
\newblock In: \emph{Proceedings of the Conf. on Computer Vision and Pattern
  Recognition}. IEEE, pp. 4612--4621.

\bibitem[{Wulfmeier et~al.(2016)Wulfmeier, Wang and
  Posner}]{wulfmeier2016watch}
Wulfmeier M, Wang DZ and Posner I (2016) Watch this: Scalable cost-function
  learning for path planning in urban environments.
\newblock In: \emph{Proceedings of the Int. Conf. on Intelligent Robots and
  Systems}. IEEE, pp. 2089--2095.

\bibitem[{Zhou et~al.(2014)Zhou, Lapedriza, Xiao, Torralba and
  Oliva}]{zhou2014learning}
Zhou B, Lapedriza A, Xiao J, Torralba A and Oliva A (2014) Learning deep
  features for scene recognition using places database.
\newblock In: \emph{Proceedings of Advances in neural information processing
  systems}. pp. 487--495.

\bibitem[{Ziebart et~al.(2008)Ziebart, Maas, Bagnell and Dey}]{ziebart-AAAI-08}
Ziebart BD, Maas AL, Bagnell JA and Dey AK (2008) Maximum entropy inverse
  reinforcement learning.
\newblock In: \emph{Proceedings of the Conf. on Artificial Intelligence},
  volume~8. AAAI, pp. 1433--1438.

\bibitem[{Ziebart et~al.(2009)Ziebart, Ratliff, Gallagher, Mertz, Peterson,
  Bagnell, Hebert, Dey and Srinivasa}]{ziebart-iros-09}
Ziebart BD, Ratliff N, Gallagher G, Mertz C, Peterson K, Bagnell JA, Hebert M,
  Dey AK and Srinivasa S (2009) Planning-based prediction for pedestrians.
\newblock In: \emph{Proceedings of the Int. Conf. on Intelligent Robots and
  Systems}. pp. 3931--3936.

\end{thebibliography}
\end{document}